\documentclass{article}
\usepackage[final]{neurips_2025}
\usepackage[utf8]{inputenc}
\usepackage[T1]{fontenc} 
\usepackage{hyperref}
\usepackage{url}
\usepackage{booktabs} 
\usepackage{amsfonts}
\usepackage{nicefrac}
\usepackage{microtype}
\usepackage[table,dvipsnames]{xcolor}

\usepackage{amsmath,amssymb}
\usepackage{wrapfig}
\usepackage{csquotes}
\usepackage{tikz}
\usepackage{subcaption}
\usepackage{adjustbox}
\usetikzlibrary{positioning, shapes.geometric, arrows.meta}
\usepackage{float}
\usepackage{colortbl}
\usepackage{multirow}

\usepackage{setspace}
\usepackage{multicol}
\usepackage{enumitem}
\usepackage{caption}
\usepackage{natbib} 
\usepackage[utf8]{inputenc} % allow utf-8 input
\usepackage[T1]{fontenc}    % use 8-bit T1 fonts
\usepackage{hyperref}       % hyperlinks
\usepackage{url}            % simple URL typesetting
\usepackage{booktabs}       % professional-quality tables
\usepackage{amsfonts}       % blackboard math symbols
\usepackage{nicefrac}       % compact symbols for 1/2, etc.
\usepackage{microtype}      % microtypography
\usepackage[utf8]{inputenc}
\usepackage{amsmath}
\usepackage{amsfonts}
\usepackage{amssymb}
\usepackage{geometry}
\usepackage{array}
\usepackage{ragged2e}
\usepackage{csquotes}
\usepackage{tikz}
\usepackage{subcaption}
\usepackage{adjustbox}
\usetikzlibrary{positioning, shapes.geometric, arrows.meta}
\usepackage{float}
\usepackage{colortbl}
\usepackage{multirow}
\usepackage{amsthm}

\theoremstyle{definition}
\newtheorem{definition}{Definition}

\newcommand{\red}[1]{\textcolor{red}{#1}}

\newcommand{\blue}[1]{\textcolor{blue}{#1}}
\usetikzlibrary{arrows.meta,positioning}
\tikzset{
  person/.style={draw, circle, inner sep=2pt, minimum size=16pt},
  edgelabel/.style={font=\ttfamily\footnotesize, fill=white, inner sep=1pt},
  >={Stealth},
  every node/.style={font=\footnotesize}
}
\colorlet{EdgePink}{Magenta} % your “pink” highlight
\title{When No Paths Lead to Rome: Benchmarking Systematic Neural Relational Reasoning}
\author{
 Anirban Das\thanks{Equal contribution. \\ Data generation code is available at~~\url{https://github.com/axd353/WhenNoPathsLeadToRome/} \\ Eval code is available at~~\url{https://github.com/erg0dic/WhenNoPathsLeadToRome/}}\\
  Cardiff University\\
  \texttt{dasa8@cardiff.ac.uk}\\
  \And
  Irtaza Khalid\footnotemark[1]\\
  Cardiff University\\
  \texttt{khalidmi@cardiff.ac.uk}\\
  \And
  Rafael Peñaloza\\
  University of Milano-Bicocca\\
  \texttt{rafael.penalozanyssen@unimib.it}\\  
  \And
  Steven Schockaert\\
  Cardiff University\\
  \texttt{schockaerts1@cardiff.ac.uk}\\ 
}

\begin{document}
\maketitle

\begin{abstract}
Designing models that can learn to reason in a systematic way is an important and long-standing challenge. In recent years, a wide range of solutions have been proposed for the specific case of systematic relational reasoning, including Neuro-Symbolic approaches, variants of the Transformer architecture, and specialised Graph Neural Networks. However, existing benchmarks for systematic relational reasoning focus on an overly simplified setting, based on the assumption that reasoning can be reduced to composing relational paths. In fact, this assumption is hard-baked into the architecture of several recent models, leading to approaches that can perform well on existing benchmarks but are difficult to generalise to other settings. To support further progress in the field of systematic relational reasoning with neural networks, we introduce NoRA, a new benchmark which adds several levels of difficulty and requires models to go beyond path-based reasoning. 
% TODO: add sentence summarising the main findings from the experiments.
\end{abstract}

\section{Introduction}
The problem of \emph{relational reasoning} involves predicting relationships between entities that are entailed from a given set of facts (expressing properties of different entities and how they are related). Entailment arises from a set of rules that a model must learn from examples. The central challenge lies in designing models capable of \emph{systematic} reasoning, a concept closely linked to compositional generalization \citep{hupkes2020compositionality}. This means that models should be able to solve test cases by applying the rules they have learned in novel ways. Recently, various neural network models have been proposed for this purpose, including neuro-symbolic approaches \citep{DBLP:conf/icml/Minervini0SGR20}, path-based methods \citep{cheng2023neural}, transformer variants \citep{edge-transformer}, and graph neural networks (GNNs) \citep{khalid2025systematic}.

Two significant problems are the lack of datasets that adequately test for systematicity, and the fact that state-of-the-art models heavily leverage the structure of existing benchmarks. CLUTRR \citep{Sinha2019CLUTRR}, the most popular benchmark for assessing systematicity, focuses on inferring family relationships. While all CLUTRR training examples can be solved in at most four inference steps, the test examples may require up to ten. Standard GNNs struggle with this kind of length generalization. Furthermore, the most successful neural methods exploit a specific characteristic of CLUTRR: the reasoning process reduces to composing relations along a single path connecting the target and source entities, where the relational facts are viewed as a knowledge graph. For example, given the path $a \xrightarrow{\textit{brother-of}} b \xrightarrow{\textit{daughter-of}} c \xrightarrow{\textit{brother-of}} d$, one can infer that $d$ is the uncle of $a$ by composing the relations \textit{brother-of}, \textit{daughter-of}, and \textit{brother-of}. We refer to this style of reasoning as \emph{path-based reasoning}.
Relational reasoning often requires going beyond path-based reasoning, but this is not reflected in existing benchmarks. The only exception is STaR \citep{khalid2025systematic}, which focuses on temporal and spatial reasoning, and requires combining the predictions of multiple relational paths. However, the main style of reasoning that is tested by this benchmark is still path-based. 

In this paper, we introduce \textbf{NoRA} (Non-Path Reasoning with Ambiguous Facts), a new benchmark which challenges state-of-the-art neural models for relational reasoning. NoRA is inspired by CLUTRR, but it intentionally breaks many of the structural assumptions in CLUTRR that state-of-the-art models are hard-coded to exploit. Like CLUTRR, the rules to be learned in NoRA are intuitive and grounded in everyday relationships—ones that humans and large language models (LLMs) naturally accept as plausible or true. However, NoRA differs from CLUTRR in three key ways.

\begin{figure}[t]
\begin{tabular}{cc}
\begin{minipage}[t]{0.4\linewidth}
\strut\vspace*{-\baselineskip}\newline
% 		\centering		
		\begin{tikzpicture}[node distance=2.2cm, every node/.style={font=\footnotesize}, ->, thick]
			\node[draw, circle, fill=red!10] (wes) at (0,2) {wes};
			\node[draw, circle] (todd) at (4,2) {todd};
			\node[draw, circle] (ann) at (2,0) {ann};
			% \node[draw, rectangle, fill=red!10] (nd) at (0,0) {no\_daughters};		
			\draw[->] (wes) -- (todd) node[midway, sloped, above] {\texttt{grandparent\_of}};
			\draw[->] (ann) -- (todd) node[midway, sloped, below] {\texttt{aunt\_of}};
			% \draw[->] (wes) -- (nd) node[pos=0.4, sloped, above] {\texttt{has\_property}};
            \draw[->] (wes) to [out=320,in=280,looseness=8] node[below] {\texttt{no\_daughters}} (wes) ;
		\end{tikzpicture}
 \end{minipage}%
&
% \begin{minipage}[t]{0.4\linewidth}		
% 		\normalsize
 \fbox{
			\begin{minipage}[t]{0.52\linewidth}            
            \strut\vspace*{-\baselineskip}\newline           
                \footnotesize
				\texttt{wes is the grandparent of todd} \\
				\texttt{wes is daughter-less} \\
				 \texttt{$\Rightarrow$ wes is the paternal grandparent of todd} \\[0.7em]
				\texttt{ann is an aunt of todd} \\
				\texttt{wes is daughter-less} \\
				\texttt{$\Rightarrow$ ann is not from wes's side of the family} \\
				\texttt{$\Rightarrow$ ann is the maternal aunt of todd}
			\end{minipage}
 }
	% \end{minipage}
 \end{tabular}   
\caption{Example where path-based reasoning fails: to derive that \texttt{ann} is \texttt{todd}'s maternal aunt, one must consider \texttt{wes}, who is not on any connecting path between \texttt{ann} and \texttt{todd}.}
\label{fig:nora_example}
\end{figure}
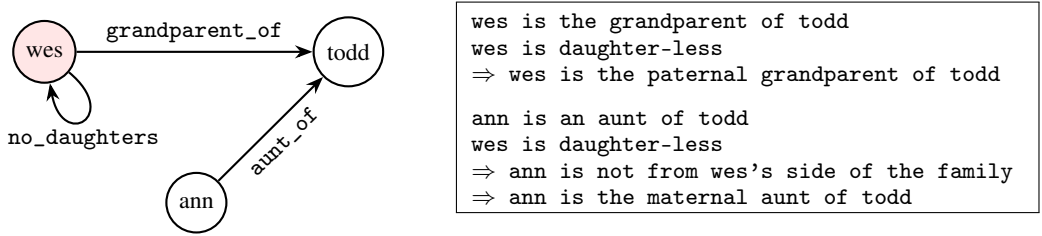

First, NoRA is specifically designed to break the path-based inductive bias that many existing relational reasoning models rely on. To this end, NoRA considers a richer set of relationships, including more fine-grained, gender-specific family roles such as \textit{maternal aunt of}, and everyday relations such as \textit{is schoolmates with} and \textit{lives in the same place as}, which often require models to go beyond path-based reasoning. Figure~\ref{fig:nora_example} illustrates such a case.
%suppose it is known that \texttt{ann} is an aunt of \texttt{todd}, \texttt{wes} is a grandparent of \texttt{todd}, and \texttt{wes} has no daughters. From this, we can infer that \texttt{wes} is the paternal grandparent of \texttt{todd}. \steven{Furthermore,} since \texttt{ann} is an aunt, and cannot be on \texttt{wes}'s side \steven{of the family} due to the absence of daughters, it follows that \texttt{ann} must be the maternal aunt of \texttt{todd}. 
In this example, we can infer that \texttt{ann} is the maternal aunt of \texttt{todd}, as explained in the figure, but to arrive at this conclusion, the reasoning must detour through the node \texttt{wes}, which is not on a path between \texttt{ann} and \texttt{todd} in the graph.  
%In later sections, we show examples of how simple instances can be systematically composed into more complex instances, paving the way for evaluating compositional generalization.

A second notable feature of NoRA is that multiple relationships may hold between a given pair of entities. These may be hierarchical (e.g.\ \texttt{ann} is both the \texttt{aunt} and the \texttt{maternal\_aunt} of \texttt{todd}) or independent (e.g.\ a person’s \texttt{brother} can also be their \texttt{schoolmate}).

Finally, NoRA incorporates a small number of ambiguous facts in its problem instances, %typically expressed as disjunctions (e.g., 
for instance expressing that $a$ is the \texttt{father\_of} either $b$ or $c$. We argue that neural relational reasoning models should be equipped to handle such ambiguity, given its ubiquity in real-world text-based reasoning. To resolve ambiguities, a model must learn to reason with constraints: the model must evaluate multiple possibilities and then (i) eliminate any possibilities that violate constraints and (ii) determine whether a given relationship holds across all the remaining possibilities.

We make the following contributions:
\begin{itemize}
	\item We introduce NoRA, a benchmark for systematic neural relational reasoning. %, which increases the difficulty of existing benchmarks, by requiring models to learn more diverse types of rules and thus go beyond path-based reasoning. We also introduce the challenge of dealing with ambiguous facts, which further complicates reasoning \steven{by requiring the model to learn to reason about constraints and to reason about multiple possibilities}.
	\item We measure the difficulty of NoRA problem instances along a number of dimensions, corresponding to the length of inference chains, the amount of ambiguity, and the extent to which the required form of reasoning goes beyond path-based reasoning. %We then construct a family of sub-datasets with varying difficulty along these dimensions.
    \item We empirically show that state-of-the-art neural models for systematic reasoning struggle on NoRA, highlighting the need for new approaches.
	%\item We release the dataset, generation code, and evaluation suite to facilitate future research in rule learning and systematic reasoning.
\end{itemize}

\section{Related work}
The problem of learning to reason has traditionally been studied in Inductive Logic Programming \citep{DBLP:journals/jlp/MuggletonR94} (ILP). Formally, given a background theory $\mathcal{B}$ and sets of positive and negative examples, ILP considers the problem of finding a set of clauses $\mathcal{H}$ such that $\mathcal{B}\cup\mathcal{H}$ logically entails every positive example and none of the negative examples. While important contributions in ILP continue to be made \citep{DBLP:journals/ml/CropperDEM22}, in recent years the focus has mostly shifted to neuro-symbolic methods, which try to solve the problem of learning to reasoning with a differentiable objective, for instance by simulating logic programming using tensor multiplication \citep{DBLP:conf/nips/YangYC17,DBLP:conf/nips/SadeghianADW19,DBLP:conf/iclr/DongMLWLZ19}, by interpreting logical connectives using fuzzy logic \citep{DBLP:journals/jair/EvansG18,DBLP:journals/jair/SourekAZSK18,DBLP:journals/ai/BadreddineGSS22}, or by using a probabilistic semantics \citep{DBLP:journals/ai/ManhaeveDKDR21}. However, these approaches are mostly designed for injecting background knowledge into the training process of a neural network model, or for making one-off predictions (e.g.\ for knowledge graph completion), rather than for systematic reasoning. 

Systematic reasoning tasks require models to learn to compose logical rules to infer conclusions. The difficulty stems from the fact that the derivations (i.e.\ the specific sequences of rule applications) that are needed for solving test instances differ from those in the training data, even if the training data contains sufficient information to learn all the required rules individually. Existing benchmarks that test for systematic reasoning include CLUTRR \citep{Sinha2019CLUTRR}, which involves predicting family relationships, GraphLog \citep{Cohen2019GraphLog}, which involves reasoning about synthetically generated knowledge graphs, and STaR \citep{khalid2025systematic}, which involves qualitative temporal and spatial reasoning. CLUTRR and GraphLog can be solved by path-based reasoning, i.e.\ the target relationship can be inferred by selecting a single relational path between the two target nodes and composing the relations along that path. STaR requires models to compose relationships among multiple paths and then taking the intersection of the resulting predictions. Reasoning is thus more involved, although still mainly path-based. Nonetheless, these benchmarks are already challenging for most approaches. Conditional Theorem Provers (CTPs) \citep{DBLP:conf/icml/Minervini0SGR20} were one of the first methods to achieve a near-perfect accuracy on CLUTRR. CTPs use a form of differentiable logic programming based on a soft unification mechanism. An important drawback of CTPs is that they are computationally expensive, which makes them impractical for many applications. Recently, a number of more efficient approaches for systematic reasoning have been proposed, such as R5 \citep{r5} and NCRL \citep{DBLP:conf/iclr/ChengAS23}. For benchmarks that only require path-based reasoning, these approaches can be  effective, but they cannot be used in more general settings such as STaR and our proposed benchmark. Edge transformers \citep{edge-transformer} are a modification of the transformer architecture, with a triangular attention mechanism that is designed to facilitate relational reasoning. They perform well on path-based benchmarks such as CLUTRR, albeit somewhat worse than CTPs, R5 and NCRL. In contrast to the aforementioned methods, their architecture does not constrain them to path-based reasoning. They also performed reasonably well on STaR. Finally, EpiGNNs \citep{khalid2025systematic} are a type of GNN model with an inductive bias for systematic relational reasoning. Their architecture is designed to support reasoning tasks where the predictions of multiple paths need to be combined, and are thus well-suited to benchmarks such as STaR.

The problem of systematic relational learning is fundamentally different from knowledge graph (KG) completion, despite the close similarities in the format of both tasks. KG completion often requires making predictions that cannot be logically entailed, by exploiting statistical biases. Because KG completion models have to capture such biases, they typically perform poorly on systematic generalization tasks. Conversely, models that are designed for systematic reasoning tend to underperform on KG completion benchmarks; see e.g.\ the comparison between NBFNet \citep{zhu2021neural} and EpiGNN by \citet{khalid2025systematic}). Interestingly, the fact that path-based reasoning is not always sufficient has also been highlighted in the context of KG completion \citep{DBLP:journals/corr/abs-2403-05130}.

% Neurosymbolic methods for learning to reason
% Existing benchmarks

%StepGame \citep{Shi2022StepGame}, STaR \citep{Khalid2024STaR}, and GraphLog \citep{Cohen2019GraphLog} 

\section{Problem setting}
Before introducing the NoRA benchmark, we introduce the problem setting and some notations.

\paragraph{Stories}
We consider the problem of reasoning about \emph{stories}, which in this context are sets of facts. % (story-facts).
Stories may contain three types of facts. First, we have binary facts, expressing a relationship between two entities, e.g.\ $\texttt{school\_mates\_with}(\texttt{ram},\texttt{irfan})$. Second, we have unary facts, 
e.g.\ $\texttt{underage}(\texttt{ryan})$, expressing a property of a single entity. 
%We will encode these unary facts using a binary relationship, where the second argument is a constant, as this will allow us to treat stories as knowledge graphs. For instance, we use $\texttt{belongs\_to}(\texttt{ryan},\texttt{underage})$ to express that the entity \texttt{ryan} has the property of being underage. Note that entities such as \texttt{ryan} are story-specific, while the constant \texttt{underage} is fixed across all stories. 
Finally, we also have facts encoding ambiguous relationships. We use the syntax of Answer Set Programming (ASP \citep{DBLP:conf/iclp/GelfondL88}) to encode such facts.\footnote{More precisely, we use the syntax of Clingo: \url{https://github.com/potassco}.}. The general form of an ambiguous fact is as follows:
\begin{align*}
l\, \{ r_1(x_1,y_1),...,r_n(x_n,y_n)\}\, u
\end{align*}
It expresses that between $l$ and $u$ of the binary facts $r_1(x_1,y_1),...,r_n(x_n,y_n)$ are true. We will specifically use such facts to encode relationships where one of the arguments is ambiguous, e.g.:
\begin{align}\label{eqExampleAmbiguousFact}
1\, \{ r(x,y_1),r(x,y_2)\} \,1
\end{align}
This encodes that $x$ is in relationship $r$ with either $y_1$ or $y_2$ (not both).
Such ambiguities often arise when reasoning about information coming from text, for instance because of ambiguous coreferences.

\paragraph{World rules}
All the stories in our dataset satisfy some regularities, which are formalized using definite rules and constraint rules. We will together refer to them as the \emph{world rules} and again use ASP syntax. \emph{Definite rules} allow us to infer relational facts from a given set of facts, e.g.: %Examples of such rules are as follows:
\begin{align*}
 \textit{living\_in\_same\_place(X,Z)} \, &\texttt{:-} \,\textit{living\_in\_same\_place(X,Y), living\_in\_same\_place(Y,Z).}\\
 %\textit{belongs\_to(X,underage)} \,&\texttt{:-} \, \textit{school\_mates\_with(X,U).}
\textit{underage(X)} \,&\texttt{:-} \, \textit{school\_mates\_with(X,U).}
\end{align*}
Uppercase arguments like $X$ denote variables. The head (left side of a rule) specifies what is inferred, while the body (right side) specifies the conditions. The first rule expresses that the \texttt{living\_in\_same\_place} relation is transitive. The second rule expresses that if somebody is school mates with somebody else, then they must be underage. \emph{Constraint rules} specify that some sets of facts can never be true at the same time. They are encoded as rules with an empty head, e.g.:
\begin{align*}
	%\texttt{:-}\, \textit{belongs\_to(X, underage), parent\_of(X, Y).}
    \texttt{:-}\, \textit{underage(X), parent\_of(X, Y).}
\end{align*}
This constraint expresses that underage people cannot be parents.

\paragraph{Answer sets}
The world rules allow us to reason about the facts that are specified in a given story $\mathcal{S}$. This process serves two purposes. First, the facts in the story are incomplete, in the sense that we can infer additional facts by applying the definite rules. Second, the constraints allow us to eliminate some of the ambiguity. To formally define the reasoning process, we need the concept of \emph{answer set}.\footnote{In general, answer sets are defined in terms of the so-called Gelfond-Lifschitz reduct \citep{DBLP:conf/iclp/GelfondL88}. For the simplified setting here, answer sets can be defined more straightforwardly.} 
For a story~$\mathcal{S}$ without ambiguity, its answer set contains all facts inferrable from~$\mathcal{S}$ via definite rules. If constraints are violated by this set,~$\mathcal{S}$ has no answer set; otherwise, $\mathcal{A} = \textit{ans}(\mathcal{S})$ denotes~$\mathcal{S}$'s answer set.
Now consider a story $\mathcal{S}$ that contains the ambiguous fact \eqref{eqExampleAmbiguousFact}. There are two possibilities: either $r(x,y_1)$ is true or $r(x,y_2)$ is true. Accordingly, we may consider two alternatives: the story $\mathcal{S}'$ in which the ambiguous fact \eqref{eqExampleAmbiguousFact} is replaced by $r(x,y_1)$, and the story $\mathcal{S}''$ in which the ambiguous fact is instead replaced by $r(x,y_2)$. We can repeat this process for all the ambiguous facts, leading to a set of unambiguous stories $\mathcal{S}_1,...,\mathcal{S}_k$. We will refer to these stories as the \emph{refinements} of $\mathcal{S}$. Then we say that $\mathcal{A}$ is an answer set of $\mathcal{S}$ if there is an  $i\in\{1,...,k\}$ such that $\mathcal{A}=\textit{ans}(\mathcal{S}_i)$. A story with ambiguous facts may thus have 0, 1 or multiple answer sets. Let us write $\textit{ref}^+(\mathcal{S})$ for the refinements of $\mathcal{S}$ which have an answer set (i.e.\ the different ways in which the ambiguities can be resolved without violating any constraints) and let $\textit{ref}^-(\mathcal{S})$ denote the other refinements (i.e.\ those where the inferred facts violate some constraints).

\paragraph{Problem formulation}
The training data consists of tuples $(\mathcal{S},x,y,\mathcal{R})$, where $\mathcal{S}$ is a story, $x$ and $y$ are entities that appear in $\mathcal{S}$, and $\mathcal{R}$ is the set of relationships that can be inferred to hold between $x$ and $y$. Formally, let us write $\textit{rels}(x,y,\mathcal{A})$ for the set of relationships that are asserted to hold between $x$ and $y$ in a given answer set $\mathcal{A}$:
\begin{align*}
\textit{rels}(x,y,\mathcal{A}) = \{r \,|\, r(x,y)\in\mathcal{A})
\end{align*}
then we have:
\begin{align*}
\mathcal{R} = \bigcap \{\textit{rels}(x,y,\mathcal{A}) \,|\, \text{$\mathcal{A}$ is an answer set of $\mathcal{S}$}\}
\end{align*}
The dataset is generated such that every story $\mathcal{S}$ has at least one answer set, i.e.\ there is always a way to resolve the ambiguities which is consistent with the constraints of the world. Test instances are queries of the form $(\mathcal{S},x,y,?)$, i.e.\ given a story and two designated entities $x$ (source) and $y$ (target), the task is to predict all relationships that can be inferred to hold between $x$ and $y$. The world rules are fixed across all training and test examples. The model is thus required to induce the world rules from the training examples, and to learn to apply them in a systematic way.

\paragraph{Example}  Suppose we have a story $\mathcal{S}$ consisting of the following facts. \footnote{More elaborate examples can be found in the Appendix \ref{seec:IntuitiveWalkthrough}.}:
\begin{align*}
&\textit{child\_of(john,mary)} &
&\textit{colleague\_of(mary,bob)} &
&\hspace{-5pt}1\{\textit{living\_in(bob,paris)},\textit{living\_in(bob,rome)}\}1\\
&\textit{living\_in(john,rome)} &
&\textit{school\_mate\_with(john,eve)} &
&\hspace{-5pt}1\{\textit{child\_of(eve,ann)},\textit{child\_of(eve,paul)}\}1
\end{align*}
There are two ambiguous facts, which means that there may be up to four answer sets. However, from $\textit{school\_mate\_with(john,eve)}$ we infer that \textit{john} is underage. We have a world rule which states that underage children live in the same place as their parents. Together with \textit{living\_in(john,rome)} we infer \textit{living\_in(mary,rome)}. We have a rule that colleagues live in the same place, allowing us to infer $\textit{living\_in(bob,rome)}$. The option \textit{living\_in(bob,paris)} is thus not consistent with the available facts (we have a constraint stating that people cannot live in two different places). The other ambiguity cannot be resolved, so the story has two answer sets. For the query $(\mathcal{S},\textit{mary},\textit{rome},?)$ the answer is thus $\mathcal{R}=\{\textit{living\_in}\}$, as \textit{living\_in(mary,rome)} is included in both answer sets. For the query $(\mathcal{S},\textit{eve},\textit{ann},?)$ the answer is $\mathcal{R}=\emptyset$, as \textit{child\_of(eve,ann)} is only included in one of the answer sets.

\section{Dataset construction}
We now present the details of our benchmark and introduce a number of metrics for measuring different aspects of problem difficulty. These difficulty measures are then used for creating systematic test splits, which will allow us to evaluate different aspects of compositional generalization.

\subsection{Data generation process}
We generate a story by randomly generating story facts. %These story facts, along with the world rules, form a logic program.
%(LP) 
%which we refer to as the story. 
We use Clingo~5.7.1 \citep{gebser2011potassco} to obtain the answer sets. An \emph{entailed atom} is an atom that appears in all the answer sets of the story but is not explicitly provided as a story fact. The entailed atoms are used to construct the queries in our benchmark. 
%Every story we generate is guaranteed to 
We only retain stories that have at least one answer set and at least one entailed atom. We generate many stories, where each story includes a different set of story facts, while the world rules remain constant across all generated stories for a dataset.  Details regarding the exact world rules, types of ambiguous facts considered, and the sampling process are in Appendix \ref{sec:data_generation}.

\subsection{Measuring problem difficulty}
We propose a number of metrics for measuring the difficulty of a given problem instance. These metrics serve two purposes. First, since we want to test for systematicity, we will consider test instances that are strictly harder than the training instances. To solve such test instances, models need to learn to compose the knowledge they have learned in novel ways (rather than learning shortcuts or memorizing computation graphs). Second, the proposed difficulty metrics will allow us to analyze model performance in a more fine-grained way.

\paragraph{Reasoning depth}
A standard notion of difficulty is the number of inference steps that are needed to infer the answer (i.e.\ the number of rule applications). Let $\mathcal{S}$ be a given story, and let $\mathcal{S}_1,...,\mathcal{S}_k$ be the refinements of $\mathcal{S}$ that are consistent with the constraints. The answer sets of $\mathcal{S}$ are then given by $\mathcal{A}_1,...,\mathcal{A}_k$ with $\mathcal{A}_i=\textit{ans}(\mathcal{S}_i)$. Let $r(a,b)$ be a fact that is included in $\mathcal{A}_i$. We define the reasoning depth of $r(a,b)$ in $\mathcal{S}_i$, written $\textit{depth}(r(a,b),\mathcal{S}_i)$, as the minimum number of inference steps that are needed to infer $r(a,b)$ from $\mathcal{S}_i$. For instance, if $r(a,b)$ is included in $\mathcal{S}_i$, we have $\textit{depth}(r(a,b),\mathcal{S}_i)=0$. Similarly, if $\mathcal{S}_i$ if a refinement that violates the constraints, we write $\textit{depth}(\bot,\mathcal{S}_i)$ for the minimum number of inference steps that are needed to establish that the constraints are violated.
The \emph{maximum reasoning depth} of a problem instance $(\mathcal{S},a,b,\mathcal{R})$ is then computed as follows:
\begin{align*}
\textit{max-depth}(\mathcal{S},a,b,\mathcal{R}) = \max\big(&\{\textit{depth}(r(a,b),\mathcal{S}_i) \,|\, \mathcal{S}_i \in \textit{ref}^+(\mathcal{S}), r\in \mathcal{R}\}\\
&\cup \{\textit{depth}(\bot,\mathcal{S}_i) \,|\, \mathcal{S}_i \in \textit{ref}^-(\mathcal{S})\}\big)
\end{align*}
The reasoning depth is determined by the hardest relation in $\mathcal{R}$ and the hardest answer set. 
%Also note that the difficulty of inferring that a given refinement $\mathcal{S}_i$ is in conflict with the constraints, is not taken into account.

\paragraph{Reasoning width}
Intuitively, the more ambiguity in a given problem instance, the harder it is to solve, all things being equal. We can straightforwardly measure the amount of ambiguity by computing the number of possible refinements of a story $\mathcal{S}$. If each ambiguous fact introduces two possibilities, then the number of possible refinements is $2^N$, with $N$ the number of ambiguous facts.\footnote{This follows because the alternatives occurring in different ambiguous facts never overlap in our dataset.} However, some ambiguous fact may not play any role in the derivation of the query, so simply counting the number of refinements may be misleading. As an alternative, we therefore focus on counting the number of unique derivations. In particular, we define the reasoning width of a fact $r(a,b)$ w.r.t.\ a story $\mathcal{S}$ as:
\begin{align*}
\textit{width}(r(a,b),\mathcal{S}) 
= &
|\{\textit{proof}(r(a,b),\mathcal{S}_i)\,|\, \mathcal{S}_i \in \textit{ref}^+(\mathcal{S}) \}| + |\{\textit{proof}(\bot,\mathcal{S}_i)\,|\, \mathcal{S}_i \in \textit{ref}^-(\mathcal{S}) \}|
\end{align*}
where we write $\textit{proof}(r(a,b),\mathcal{S}_i)$ for the derivation which proves that $r(a,b)$ can be derived from $\mathcal{S}_i$. If there are multiple proofs, we fix $\textit{proof}(r(a,b),\mathcal{S}_i)$ to be the shortest one, with ties broken arbitrarily. Similarly, $\textit{proof}(\bot,\mathcal{S}_i)$ denotes a minimal proof that $\mathcal{S}_i$ violates the constraints. In other words, the width of $r(a,b)$ is the sum of the number of distinct derivations of $r(a,b)$, across all the answer sets of $\mathcal{S}$, and the  number of distinct derivations of constraint violation, across all refinements of $\mathcal{S}$ without an answer set. The maximal width of a problem instance $(\mathcal{S},a,b,\mathcal{R})$ is then computed as the reasoning width for the hardest relation in $\mathcal{R}$:
\begin{align*}
\textit{max-width}(\mathcal{S},a,b,\mathcal{R}) = \max\{\textit{width}(r(a,b),\mathcal{S}) \,|\,  r\in \mathcal{R}\}
\end{align*}

% Ambiguity introduces a new notion of difficulty. For the entailed atom \texttt{living\_in(ryan, kgp)}, Figure~\ref{fig:kanr_ambiguous_fig}-b shows eight derivation branches (i–viii), of which branches v–viii lead to contradictions and share the same structure. Among the positive branches, branches i and ii yield identical derivations, as do iii and iv. 

% We define the \textbf{number of unique branches} for a query as the sum of:
% \begin{itemize}
% 	\item the number of distinct derivations that yield the entailed atom across all consistent stable models, and
% 	\item the number of distinct derivations that lead to contradiction in the remaining branches.
% \end{itemize}

% For the example in Figure~\ref{fig:kanr_ambiguous_fig} (with story facts in 2a and the entailed atom \texttt{living\_in(ryan, kgp)}), this number is 3. Details of how we encode the examples with ambiguous facts as graphs are in supplementary. We now have two notions of difficulty:
% \begin{itemize}
% 	\item \textbf{Max Reasoning Depth:} the maximum number of inference steps required to derive the entailed atom in any consistent branch.
% 	\item \textbf{Number of Unique Branches:} the number of alternative derivations—both successful and failed—that must be considered to correctly infer the entailed atom.
% \end{itemize}

\paragraph{Non-path reasoning}
In the case of CLUTRR, all the required rules are of the following form 
\begin{align}\label{eqCLUTRRrule}
r(X,Z) \texttt{:-} r_1(X,Y),r_2(Y,Z)
\end{align}
If all rules are like this, then the problem of inferring a relational fact $s(a,b)$ boils down to (i) finding an informative path connecting $a$ and $b$ (where we view the facts as the edges of a knowledge graph), and (ii) repeatedly applying rules of the form \eqref{eqCLUTRRrule} to replace two adjacent edges by a single edge (representing the composition of the two given relations), until we end up with a single edge connecting $a$ and $b$.\footnote{See \citep{khalid2025systematic} for a formal proof of this claim.} Many approaches for systematic relational reasoning are closely aligned with this idea.
%view of relational reasoning. 
As such, problem instances that require going beyond this kind of path-based reasoning can be expected to present difficulties for many models. We introduce two metrics to measure the extent to which a problem instance requires going beyond path-based reasoning.

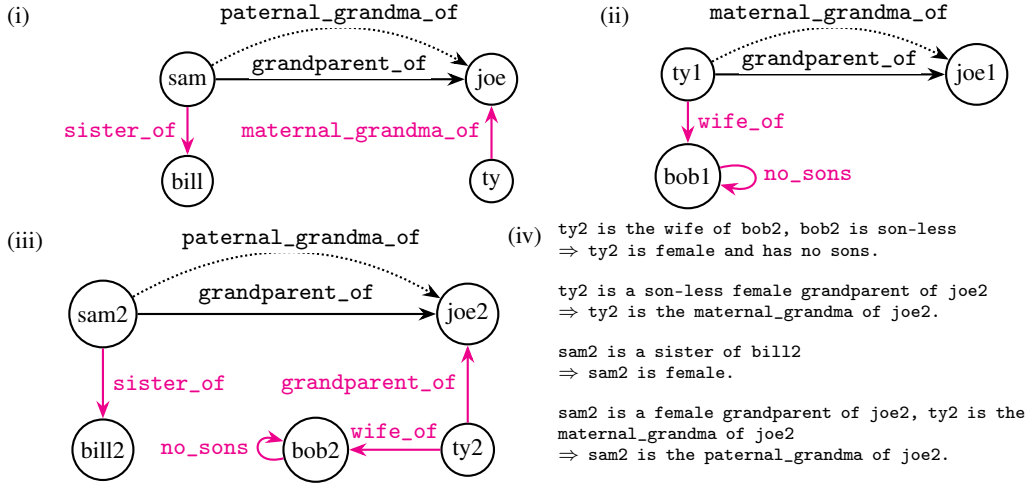
\begin{figure}[t]
\begin{minipage}[t]{0.49\linewidth}
\centering
\begin{tikzpicture}[thick,->,x=1cm,y=0.9cm,baseline=(current bounding box.north)]
  % Nodes
  \node[person] (bill) at (-1.5,-1.5) {bill};
  \node[person] (sam)  at (-1.5,0) {sam};
  \node[person] (ty)   at (2.5,-1.5) {ty};
  \node[person] (joe)  at ( 2.5,0) {joe};

  % Edges
  \draw (sam) -- (joe) node[midway, above, edgelabel]{\texttt{grandparent\_of}};
  \draw[EdgePink] (sam) -- (bill) node[midway,left]{\texttt{sister\_of}};
  \draw[densely dotted, bend left=30] (sam) to
       node[above, edgelabel]{\texttt{paternal\_grandma\_of}} (joe);
  \draw[EdgePink,] (ty) to
       node[midway, left]{\texttt{maternal\_grandma\_of}} (joe);

  % Label to the LEFT at the top
  \node[anchor=east] at ([xshift=-0.2cm,yshift=-0.55em]current bounding box.north west) {\footnotesize (i)};
\end{tikzpicture}
\end{minipage}
\hfill
\begin{minipage}[t]{0.49\linewidth}
\centering
\begin{tikzpicture}[thick,->,x=1cm,y=0.9cm,baseline=(current bounding box.north)]
  % Nodes
  \node[person] (ty1)  at (-1.8,0) {ty1};
  \node[person] (joe1) at ( 2,0) {joe1};
  \node[person] (bob1) at ( -1.8,-1.5) {bob1};

  % Edges
  \draw (ty1) -- (joe1) node[midway, above, edgelabel]{\texttt{grandparent\_of}};
  \draw[densely dotted, bend left=28] (ty1) to
       node[midway, above, edgelabel]{\texttt{maternal\_grandma\_of}} (joe1);
  \draw[EdgePink] (ty1) -- (bob1) node[midway, right]{\texttt{wife\_of}};
  \draw[EdgePink] (bob1) to[loop right, looseness=7]
        node[right]{\texttt{no\_sons}} (bob1);

  % Label to the LEFT at the top
  \node[anchor=east] at ([xshift=-0.2cm, yshift=-0.55em]current bounding box.north west) {\footnotesize (ii)};
\end{tikzpicture}
\end{minipage}

\vspace{0.4em} % small gap between rows

% ---------- BOTTOM ROW ----------
\begin{minipage}[t]{0.46\linewidth}
\centering
\begin{tikzpicture}[thick,->,x=1cm,y=0.9cm,baseline=(current bounding box.north)]
  % Nodes
  \node[person] (bill2) at (-2.8,-2) {bill2};
  \node[person] (sam2)  at (-2.8,0)      {sam2};
  \node[person] (ty2)   at (2,-2) {ty2};
  \node[person] (joe2)  at (2,0)    {joe2};
  \node[person] (bob2)  at ( 0,-2){bob2};

  % Edges
  \draw[EdgePink] (ty2) -- (joe2) node[midway, left]{\texttt{grandparent\_of}};
  \draw[EdgePink] (ty2) -- (bob2) node[midway, above]{\texttt{wife\_of}};
  \draw[EdgePink] (bob2) to[loop left, looseness=5]
        node[left]{\texttt{no\_sons}} (bob2);
  \draw (sam2) -- (joe2) node[midway, above]{\texttt{grandparent\_of}};
  \draw[densely dotted, bend left=30] (sam2) to
       node[pos=0.55, above, edgelabel]{\texttt{paternal\_grandma\_of}} (joe2);
  \draw[EdgePink] (sam2) -- (bill2) node[midway, right]{\texttt{sister\_of}};

  % Label to the LEFT at the top
  \node[anchor=east] at ([xshift=-0.2cm,yshift=-0.55em]current bounding box.north west) {\footnotesize (iii)};
\end{tikzpicture}
\end{minipage}
\hfill
\begin{minipage}[t]{0.52\linewidth}
% Wrap the derivation in a TikZ node so the left label can align with its TOP
\begin{tikzpicture}[baseline=(box.north), every node/.style={inner sep=0}]
  % The text block
  \node[anchor=north west, text width=\linewidth] (box) {%
    {\scriptsize\ttfamily
    ty2 is the wife of bob2, 
    bob2 is son-less \\
    $\Rightarrow$ ty2 is female and has no sons.\\[0.95em]
    ty2 is a son-less female grandparent of joe2\\
    $\Rightarrow$ ty2 is the maternal\_grandma of joe2.\\[0.95em]
    sam2 is a sister of bill2\\
    $\Rightarrow$ sam2 is female.\\[0.95em]
    sam2 is a female grandparent of joe2, 
    ty2 is the maternal\_grandma of joe2\\ 
    $\Rightarrow$ sam2 is  the paternal\_grandma of joe2.\\[0.95em]
    \rmfamily}}
  ;
  % Label to the LEFT at the top of the text block
  \node[anchor=east] at ([xshift=-0.2cm,yshift=-0.55em]box.north west) {\footnotesize (iv)};
\end{tikzpicture}
\end{minipage}  
\caption{Source entities are \emph{sam}, \emph{ty1}, and \emph{sam2}, while target entities are \emph{joe}, \emph{joe1}, and \emph{joe2} for the queries accompanying stories (i), (ii) and (iii), respectively. Solid edges represent the relationships explicitly in the story. Dashed edges are entailed relationships between source–target pairs. Pink edges indicate edges that do not lie on any path between the source and target. Panel (iv) illustrates a derivation of the entailed fact in story (iii). It uses all four off-path edges, hence the query from story (iii) has an OPEC value of 4. The queries in stories (i) and (ii) each have an OPEC value of 2.}
\label{fig:opec_compositional}
\end{figure}

% Thus, even though \texttt{ann} and \texttt{todd} are connected through paths in the graph, determining that \texttt{ann} is the \texttt{maternal\_aunt} of \texttt{todd} requires reasoning that detours through \texttt{wes}—an entity that lies off the direct paths between them. While the precise world rules enabling this derivation are drawn from the \textbf{KANR-full} universe, they closely mirror real-world intuitions, and we expect the reasoning to be readily followable without needing to reference the rules explicitly.  This type of reasoning is not captured by standard path-based models, making KANR a uniquely challenging benchmark for evaluating generalization beyond structural biases.
% Few more examples are in Figure~\ref{fig:kanr_nonpath_fig}-b-e. 

% To quantify the reasoning complexity of KANR examples that go beyond path-based inference, we introduce two diagnostic metrics based on the structure of derivations:

The first metric, \textbf{backtrack load (BL),} is based on the observation that for path-based derivations, the number of inference steps is always one less than the number of entities involved in the derivation. In contrast, for more complex derivations, we often see a higher number of inference steps, relative to the number of entities. We thus define $\emph{BL}(\tau)$ for a derivation $\tau$ as the ratio of the number of inference steps to the number of entities involved. The maximum backtrack load of a problem instance is then: % given by:
\begin{align*}
\textit{max-BL}(\mathcal{S},a,b,\mathcal{R}) = \max\{\textit{BL}(\textit{proof}(r(a,b),\mathcal{S}_i)) \,|\, \mathcal{S}_i \in \textit{ref}^+(\mathcal{S}), r\in \mathcal{R}\}
\end{align*}
%
%It captures the density of inference within an example. For each derivation branch, we compute the number of inference steps used and divide it by the number of distinct entities that occur in the derivation (excluding special constants such as \texttt{underage}, \texttt{female}, \texttt{no\_sons}, etc.). We take the \emph{maximum} value across all branches as the \textbf{Backtrack Load} for the example. Intuitively, a higher BL indicates that a small number of entities are involved in a relatively large number of logical steps—reflecting back-and-forth reasoning.
%	
The second metric is called \textbf{off-path edge count (OPEC)}. For a given derivation $\tau$ of a fact $r(a,b)$, we define $\emph{OPEC}(\tau)$ as the number of edges that appear in $\tau$ which are not on any direct path between $a$ and $b$ (if we view relational facts as the edges of a knowledge graph). We then define the maximal OPEC of a problem instance as:
\begin{align*}
\textit{max-OPEC}(\mathcal{S},a,b,\mathcal{R}) = \max\{\textit{OPEC}(\textit{proof}(r(a,b),\mathcal{S}_i)) \,|\, \mathcal{S}_i \in \textit{ref}^+(\mathcal{S}), r\in \mathcal{R}\}
\end{align*}
We drop the prefix max- and refer to these objects as BL and OPEC. Figure~\ref{fig:opec_compositional} illustrates how OPEC measures the extent of non-path reasoning. 
%During training, models encounter examples like (b--d) with small OPEC, while (e) is constructed by stitching together smaller examples into one with a higher OPEC value. This enables evaluating systematic generalization to harder cases. Examples with even higher OPEC values appear in the supplementary material.

% However, the metric depends on how inference steps are defined, and can be noisy—assigning high values even in cases where no true backtracking occurs, either on-path or off-path. As a result, BL is prone to false positives. That said, it is capable of detecting complex reasoning that still occurs along the path between the source and target. 
These two metrics are complementary. BL captures whether the reasoning process needs to go back-and-forth along a given relational path. This back-and-forth reasoning is often required for the problems in our benchmark, even when all the edges involved are on a single path between the two query entities. We expect this to be challenging for many approaches, especially methods such as NCRL and R5 which by design only make a single pass over a given path.
OPEC captures whether any off-path reasoning is required. Path-based models typically ignore any edges that are not on a direct path between the two query entities. For more  discussion see Appendix \ref{sec:bl_vs_opec}.

%*******
\subsection{Training distribution and held-out test sets}

%\begin{figure}[t]
	%\centering
	%\includegraphics[width=0.85\textwidth]{distribution_of_properties.png} 
	%\caption{\steven{Distribution of problem difficulty in the training data.}}
	%\label{fig:kanr_training_fig}
%\end{figure}
\begin{wraptable}{r}{0.5\textwidth}
% \vspace{-8ex}
	\caption{Overview of the dataset splits. Values that require the model to generalize from the training distributions are highlighted in red.}
	\label{tab:kanr_test_sets}
    \footnotesize
    \centering
    \setlength\tabcolsep{3pt}
    \centering
	\begin{tabular}{lcccc}
		\toprule
		\textbf{Name}  &  \textbf{Depth} &  \textbf{Width} &  \textbf{BL}  &  \textbf{OPEC}\\
		\midrule
        \textbf{Train-a}  & $\leq 6$ & $\leq 5$ & $\leq 1.5$ & $\leq 2$\\
        \textbf{Train-\blue{na}} & $\leq 6$ & \textbf{\blue{1}} & $\leq 1.5$ & $\leq 2$\\
        \midrule
		\textbf{Test-\red{D}} &  \red{$\mathbf{> 6}$} & $\leq 5$ & $\leq 1.5$ & $\leq 2$ \\
        \textbf{Test-\red{W}} &  $\leq 6$ & \red{$\mathbf{> 5}$} & $\leq 1.5$ & $\leq 2$ \\
	    \textbf{Test-\red{BL}} &  $\leq 6$ & $\leq 5$ & \red{$\mathbf{> 1.5}$} & - \\
		\textbf{Test-\red{OPEC}} &  - & - & - & \red{$\mathbf{\geq 3}$} \\	
        \textbf{Test-In-dist} & $\leq 6$ & $\leq 5$ & $\leq 1.5$ & $\leq 2$\\
        \midrule
        \textbf{Test-\red{D}-\blue{na}} &  \red{$\mathbf{> 6}$} & \textbf{\blue{1}} & $\leq 1.5$ & $\leq 2$ \\        
	    \textbf{Test-\red{BL}-\blue{na}} &  $\leq 6$ & \textbf{\blue{1}} & \red{$\mathbf{> 1.5}$} & - \\
		\textbf{Test-\red{OPEC}-\blue{na}} &  - & \textbf{\blue{1}} & - & \red{$\mathbf{\geq  3}$} \\		
        \textbf{Test-In-dist-\blue{na}} & $\leq 6$ &  \textbf{\blue{1}} & $\leq 1.5$ & $\leq 2$\\
		\bottomrule
	\end{tabular}	
\end{wraptable}
The training set for NoRA contains examples whose difficulty, according to the four proposed metrics, is controlled: reasoning depth $\leq 6$, reasoning width $\leq 5$, BL $\leq 1.5$ and OPEC $\leq 2$. The marginal distribution of these four metrics within the training dataset covers a  variety of examples (Appendix \ref{fig:training_distributions}), which is essential for enabling models to generalize systematically.
%Figure \ref{fig:kanr_training_fig} shows a histogram of the difficulty levels of the training examples.
We have also created a separate training set which is free of ambiguity, i.e.\ where all examples have reasoning width 1. To rigorously test generalization, we define several held-out evaluation subsets, each focused on specific types of reasoning that go beyond what the model encounters during training. We have four such out-of-distribution test sets involving ambiguities and three which do not. Each of these out-of-distribution test sets extends the difficulty level of the problem instances according to one of the considered difficulty metrics. Finally, we also created in-distribution test sets, containing unseen problem instances with similar characteristics as those from the training set. An overview of the different datasets is shown in Table \ref{tab:kanr_test_sets}.

\section{Experiments}

We evaluate a number of state-of-the-art models on NoRA. Pure path-based methods, such as NCRL and R5, are limited to path-based inference by design, and are thus not suitable. CTPs are too inefficient to handle the large number of rules that needs to be learned for NoRA, and they cannot model constraints. We therefore focus our analysis on the following methods. \textbf{Edge Transformers} (ETs) \citep{edge-transformer} are more versatile than other methods for systematic reasoning, and thus a natural candidate for the more challenging setting presented by NoRA. However, they cannot naturally model multiple relationships between the same entities (i.e.\ the edge index cannot have degeneracies). We therefore consider two versions of ETs: a vanilla ET, where a single relationship is chosen for each entity pair, arbitrarily, and others are simply ignored (single-edge) and a modified ET in which the edge embeddings are averaged if there are multiple relationships (multi-edge). We also evaluate transformers with \textbf{relation-aware self attention} (RAT) \citep{DBLP:conf/naacl/ShawUV18}, as a precursor to ETs. Next, we evaluate \textbf{EpiGNNs} \citep{khalid2025systematic}, which are the state-of-the-art on STaR (the only existing benchmark that goes beyond path-based systematic reasoning). We consider two variants: one with the original margin loss %, designed for single-label link prediction, 
and one with a binary cross-entropy loss, with the latter intuitively being more suitable for the multi-label setting. We consider both minimum and multiplication for aggregation. Finally, we evaluate \textbf{NBFNet} \citep{zhu2021neural} and \textbf{R-GCNs} \citep{DBLP:conf/esws/SchlichtkrullKB18} as representative GNN models. To evaluate these models, we encode ambiguities in the graph representation of stories using special edges (Appendix \ref{sec:AmbFacts}).

% \begin{table}
% \footnotesize
% % \scriptsize
% \centering
% %\setlength\tabcolsep{3pt}
% \caption{Results of state-of-the-art models for systematic reasoning on the NoRA test sets (accuracy).\label{tabMainResults}}
% \begin{tabular}{lccccccc}
% \toprule
% & \multicolumn{4}{c}{\textbf{Trained with ambiguity}} & \multicolumn{3}{c}{\textbf{Trained without ambiguity}}\\
% \cmidrule(lr){2-5}\cmidrule(lr){6-8}
% & \textbf{D} & \textbf{W} & \textbf{BL} & \textbf{OPEC} & \textbf{D-na} & \textbf{BL-na} & \textbf{OPEC-na}\\
% \midrule
% ET (single-edge) & 0.734 & 0.489 & \textbf{0.791} & \textbf{0.786} & 0.062 & 0.494 & 0.056 \\
% ET (multi-edge) & 0.781 & \textbf{0.739} & 0.703 & 0.245 & 0.104 & \textbf{0.822} & \textbf{0.104} \\
% RAT (single-edge) & \textbf{0.812} & 0.676 & 0.668 & 0.540 & 0.021 & 0.768 & 0.023 \\
% RAT (multi-edge) & 0.656 & 0.490 & 0.615 & 0.234 & 0.181 & 0.493 & 0.092 \\
% EpiGNN-\texttt{min} (margin) & 0.491& 0.176 & 0.000 & 0.000 & 0.485 & 0.000 & 0.000 \\
% EpiGNN-\texttt{min} (BCE) & 0.495 & 0.445 & 0.131 & 0.021 & 0.488 & 0.022 & 0.040 \\
% EpiGNN-\texttt{mul} (BCE) & 0.686 & 0.501 & 0.156 & 0.010 & 0.762 & 0.027 & 0.046 \\
% NBFNet (margin) & 0.000 & 0.000 & 0.000 & 0.000 & 0.000 & 0.000 & 0.000 \\
% NBFNet (BCE) & 0.531 & 0.460 & 0.153 & 0.009 & \textbf{0.764} & 0.012 & 0.043 \\
% R-GCN & 0.672 & 0.283 & 0.051 & 0.032 & 0.740 & 0.018 & 0.012 \\
% \bottomrule
% \end{tabular}
% \end{table}

\begin{table}[t]
% \footnotesize
\scriptsize
\centering
\caption{Results of state-of-the-art models for systematic reasoning on the NoRA test sets.\label{tabMainResults}}
\begin{tabular}{llccccccccc}
\toprule
& & \multicolumn{5}{c}{\textbf{Trained with ambiguity}} & \multicolumn{4}{c}{\textbf{Trained without ambiguity}}\\
\cmidrule(lr){3-7}\cmidrule(lr){8-11}
& & \textbf{In-dist} & \textbf{D} & \textbf{W} & \textbf{BL} & \textbf{OPEC} & \textbf{In-dist-na} & \textbf{D-na} & \textbf{BL-na} & \textbf{OPEC-na}\\
\midrule
\multirow{10}*{\rotatebox{90}{\textbf{Exact-match Accuracy}}}
& ET (single-edge) & 0.885 & \textbf{0.741} & 0.703 & 0.245 & \textbf{0.060} & \textbf{0.800} & \textbf{0.822} & \textbf{0.104} & \textbf{0.110} \\
& ET (multi-edge) & \textbf{0.900} & 0.493 & \textbf{0.790} & \textbf{0.785} & 0.037 & \textbf{0.800} & 0.494 & 0.056 & 0.077 \\
& RAT (single-edge) & 0.721 & 0.494 & 0.615 & 0.234 & 0.042 & 0.800 & 0.493 & 0.092 & 0.094 \\
& RAT (multi-edge) & \textbf{0.900} & 0.676 & 0.668 & 0.540 & 0.028 & 0.827 & 0.768 & 0.023 & 0.017 \\
& EpiGNN-\texttt{min} (margin) & 0.334 & 0.491& 0.176 & 0.000 & 0.000 & 0.208 & 0.485 & 0.000 & 0.000 \\
& EpiGNN-\texttt{min} (BCE) & 0.451 & 0.665 & 0.456 & 0.154 & 0.005 & 0.475 & 0.488 & 0.008 & 0.025 \\
& EpiGNN-\texttt{mul} (BCE) & 0.520 & 0.604 & 0.491 & 0.156 & 0.009 & 0.539 & 0.716 & 0.027 & 0.045 \\
&NBFNet (margin) & 0.000 & 0.000 & 0.000 & 0.000 & 0.000 & 0.000 & 0.000 & 0.000 & 0.000 \\
&NBFNet (BCE) & 0.576 & 0.531 & 0.460 & 0.153 & 0.009 & 0.679 & 0.764 & 0.012 & 0.043 \\
&R-GCN & 0.347 & 0.672 & 0.283 & 0.051 & 0.032 & 0.579 & 0.740 & 0.018 & 0.012 \\
\bottomrule
\end{tabular}
\end{table}
\paragraph{Main results} The results are shown in Table \ref{tabMainResults}, in terms of exact-match accuracy (i.e.\ we measure if the model's prediction of the relation set $\mathcal{R}$ exactly matches the ground truth). Models trained on \emph{Train-a} are evaluated on the test sets with ambiguity, while models trained on \emph{Train-na} are evaluated on the remaining test sets. 
ETs emerge as the best-performing model. %, with the D-na test set as the only exception. 
All models perform poorly on OPEC, BL-na and OPEC-na.  Surprisingly, for most models, performance on test-W is reasonable. Furthermore, all models perform better on BL than on BL-na, despite the fact that BL was assumed to be harder.
Further analysis has shown that models are exploiting shortcuts to solve the majority of ambiguous problems (see Appendix \ref{appDiagnosingAppendixPerformance}). The GNN methods all perform poorly on the BL and OPEC test sets, which can be explained by their strong alignment with path-based reasoning. In fact, the GNN models are even performing poorly on the in-distribution test sets, for the same reason. Among the GNN models, EpiGNNs with BCE loss and multiplication-based aggregation perform better. The results also confirm that the margin-based loss is unsuitable for the multi-label setting.

\begin{figure}[t]
    \centering
    \includegraphics[width=1.\linewidth]{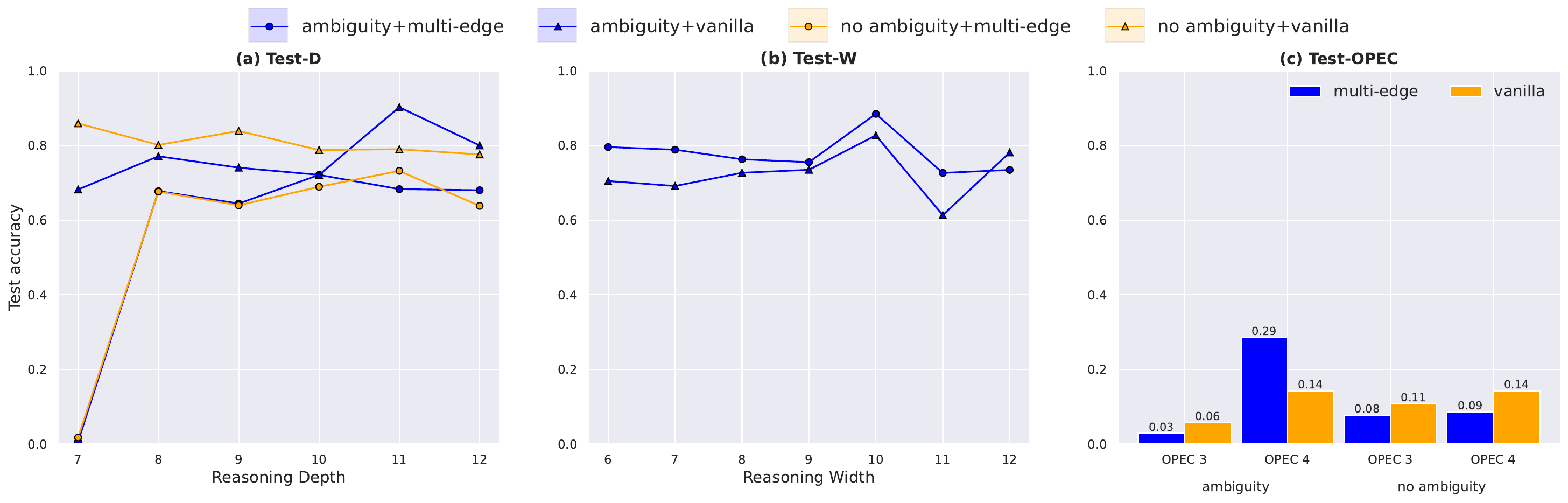}
    \caption{Analysis of the performance of ETs on various splits of the dataset. 
    % Macro F1 scores with equal weights per class for F1 computation highlights class imbalances in the test splits. 
    }
    \label{fig:depth-first-et-2}
\end{figure}

\begin{figure}[t]
    \centering        
    % \begin{subfigure}[b]{0.24\textwidth}
    %     \centering
    %     \includegraphics[width=1.1\textwidth]{opec-et.pdf}
    %     \vspace{-2ex}        
    %     % \caption{Performance of ETs on Test-OPEC\label{fig:opec-edget}}
    %     \caption{\label{fig:opec-edget}}
    % \end{subfigure}
    % \hfill
    \begin{subfigure}[b]{0.38\textwidth}
        \centering        
        \includegraphics[width=\textwidth]{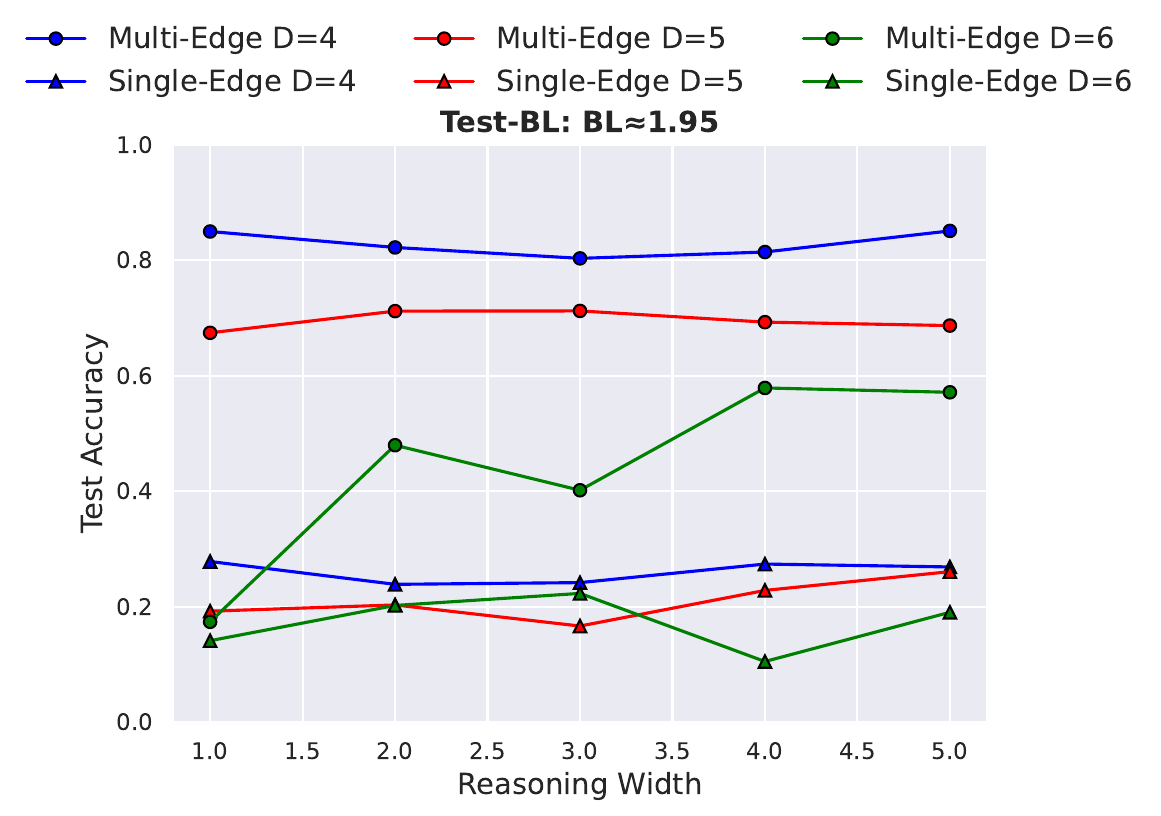}
        % \caption{Performance of ETs on Test-D\label{fig:bl-test-d-edget}}
        \caption{\label{fig:bl-test-d-edget}}
    \end{subfigure}
    \hfill
    \begin{subfigure}[b]{0.3\textwidth}
        \centering
        \includegraphics[width=\textwidth]{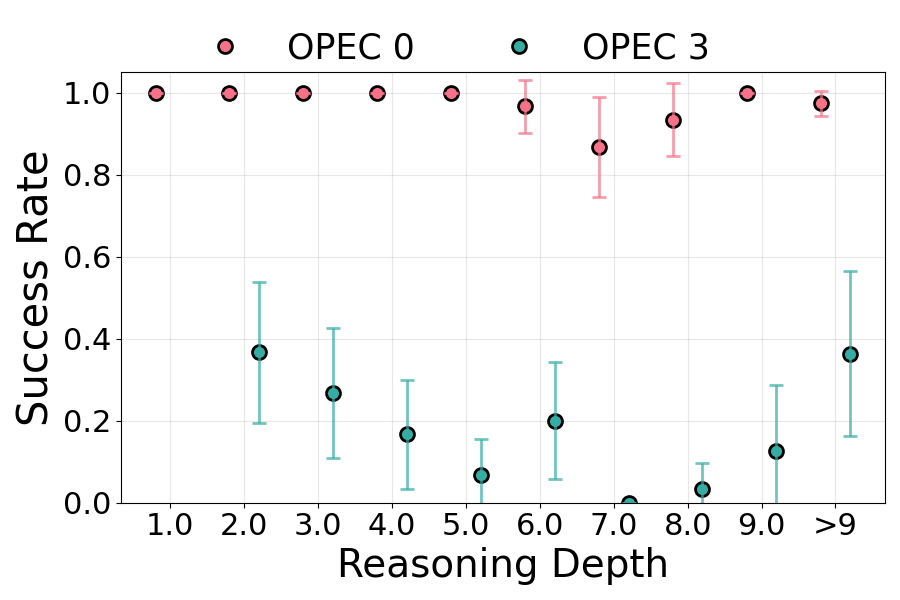}
         % \caption{Analysis of o3\label{fig:opec_query}}
        \caption{\label{fig:opec_query}}
    \end{subfigure}
    \hfill
    \begin{subfigure}[b]{0.3\textwidth}
        \centering
        \includegraphics[width=\textwidth]{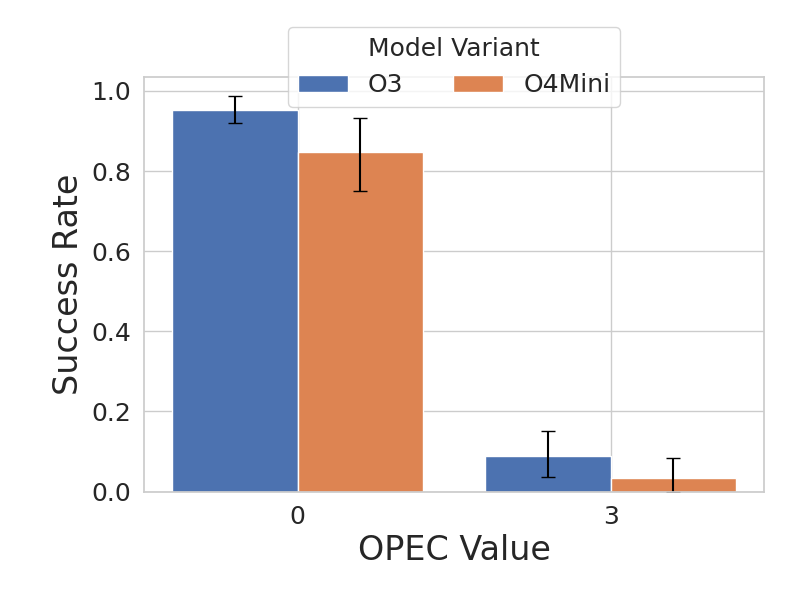}
         % \caption{Comparison of o3 and o4-mini\label{fig:o3_vs_o4}}
        \caption{\label{fig:o3_vs_o4}}
    \end{subfigure}
    \caption{
    %Performance comparisons of different models and OPEC values
    (a) Breakdown of the performance of edge transformers on Test-D; (b) analysis of o3 on non-ambiguous stories; (c) a comparison between o3 and o4-mini on non-ambiguous stories.}
    \label{fig:opec_comparisons}
\end{figure}

\paragraph{Analysis of ET performance} 
Figure \ref{fig:depth-first-et-2} breaks down the performance of the Edge Transformer on test-D, test-W and test-OPEC. Surprisingly, the performance decline is minimal along the considered difficulty axes. For instance, in the case of Test-D, the results for reasoning depth 12 are almost as good as those for reasoning depth 7, for the vanilla model. Similarly, apart from the dip at depth 7, the multi-edge model performs similarly between depths 8 to 12. However, these results have to be interpreted with caution. Recall that the test problems were obtained by random sampling. Obtaining hard instances in this way is difficult, meaning that we cannot easily test how the model would perform when the reasoning depth is higher than 12 or OPEC is higher than 4, for instance. This is something that we have addressed by introducing a variant of our benchmark, called NoRA v1.1, as explained below. Another consequence of the fact that randomly sampled problem instances are rarely hard relates to the correlations between the difficulty metrics. For instance, a problem with high reasoning depth will typically have low OPEC, and a problem with high OPEC will typically have low reasoning depth. Problems with high reasoning depth may thus be solved well because they are easier in other respects, rather than because the generalization abilities of the model. We analyze this in Figure \ref{fig:bl-test-d-edget}, where we show the performance for different reasoning depths, \emph{while controlling for both BL and reasoning width}. In this case, we can see a dramatic decline in performance when going from reasoning depth 4 (where the multi-edge ET achieves accuracies above 0.8) to reasoning depth 6 (where the performance varies from around 0.2 to 0.6). Interestingly, in this analysis, the multi-edge variant also clearly outperforms the single-edge variant. This reflects the need for more informationally complete input representations for problems with higher BL.

%especially for higher BL, when the number of inference hops is larger for a given set of involved nodes. 

% We see an adaptation with respect to the test splits for ET: depth performance is higher for single-edge ETs whilst width performance is higher for multi-edge ETs.

% We can see a very clear drop in performance between OPEC 3 and 4. While 

% Figure \ref{fig:bl-test-d-edget}

% \todo{Briefly discuss breakdown in Figure \ref{fig:opec_comparisons}.}

\begin{figure}[t]
    \centering        
    \includegraphics[width=\textwidth]{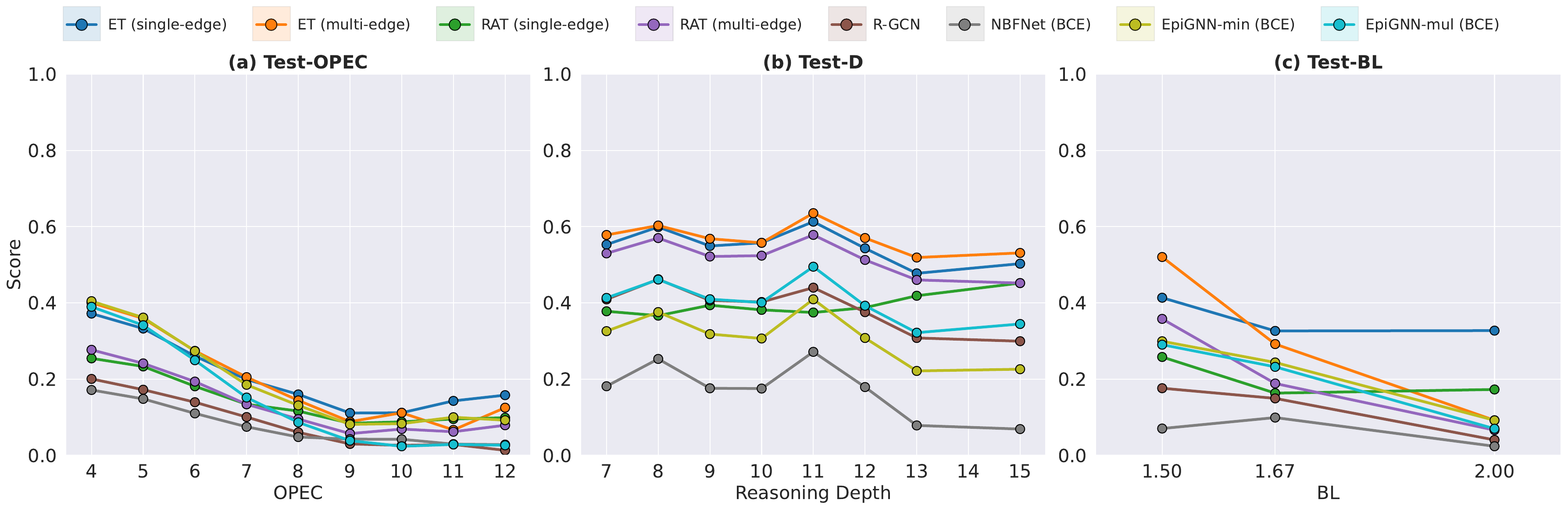}
    \caption{Results for the expanded version of NoRA (v1.1) that uses recursive subgraph expansion to generate harder splits along the axes: (a) OPEC, (b) Reasoning Depth (c) BL. }
    \label{fig:nora11}
\end{figure}

\paragraph{Evaluating Large Reasoning Models}
NoRA was designed to test the compositional generalization abilities of neural systematic reasoning models. The defined NoRA world rules are realistic, as most humans and large language models would deem them true or likely (see Appendix \ref{sec:NoRARealWorldRuleBase}). This is a desirable property for evaluating the systematic generalization and rule learning capabilities of Large Reasoning Models (LRMs) \citep{zhu2023large}. We also evaluated the LRMs o3 and o4-mini on a subset of NoRA problems, \emph{when explicitly given the entire set of world rules} (only in the LRM experiments are the rules explicitly provided; in all other settings, they must be induced by the model). Being able to apply the correct rules is clearly a prerequisite for solving the considered learning tasks. For this experiment (details in Appendix \ref{sec:lrmexp}), we only consider problem instances without ambiguity, as we want to focus on the extent to which these models can deal with off-path reasoning, and we only consider problem instances where there is a single best label. We provide the models with two in-context demonstrations.
%We design a query completion task using unambiguous data where each query has exactly one correct answer. The world rules are provided, and this task specifically measures how LLMs handle non-path-based reasoning cases. Given \textbf{world rules}, story facts, and a query (entity pair), the model must predict the correct single relationship. 
Success is measured by exact match with the ground truth label. 
The results for o3 are shown in Figure \ref{fig:opec_query}. While the model achieves near-perfect accuracy for problem instances with OPEC 0, the performance drops dramatically for OPEC 3, where none of the problem instances of reasoning depth 7 were answered correctly. Even when the world rules are explicitly provided—and are rules the LRM is already familiar with through pretraining—the model fails to apply them correctly to problem instances, highlighting the inherent difficulty of off-path reasoning tasks. Surprisingly, for higher inference depths, the performance is slightly better. This is due to the presence of instances where the LRM can apply shortcuts (Appendix \ref{sec:LLRshortcuts}). As shown in Figure \ref{fig:o3_vs_o4}, the performance of o4-mini is slightly worse than that of o3.
%We evaluate the performance of \O3 and O4-mini models on this task, both of which struggle with high-OPEC examples (see Fig.~\ref{fig:opec_comparison}). 
This non-path-based reasoning analysis aligns with findings from \citet{dziri2023faith} on pure reasoning tasks. As an auxiliary task, we tested the model's ability to recover the necessary world rules for solving the task (Appendix \ref{sec:lrmexp}).

\paragraph{Additional datasets: NoRA v1.1 and HetioNet}
To further support future work on neural relational reasoning, we introduce two additional datasets. First, we introduce a variant of NoRA, called NoRA v1.1, where problem instances are sampled in a more systematic way, using the recursive subgraph expansion technique \citep{khalid2025systematic}. This has two consequences. First, it means that we can easily create problem instances with larger OPEC, reasoning depth, and BL values. As a result, we can include examples with higher structural complexity in the training set (e.g.\ allowing OPEC values up to 3), and include much harder problem instances in the test sets. Second, by generating the problem instances in this way, we can guarantee that every test example can be obtained by a \emph{stitching together} process of multiple training examples. As a result, we are guaranteed that a model which achieves compositional generalization can solve every test instance.
% This enables the construction of a new dataset, \textbf{NoRA v1.1} (Appendix~\ref{sec:NoRA1.1}), which includes:  
% 1) Training examples with higher structural complexity (e.g., the \texttt{Train-a} split contains only examples with OPEC~$\leq$~3); and  
% 2) Test examples in all three splits that are generated by \emph{stitching together} process multiple training examples, with no ambiguous stories included.
To illustrate this stitching together process, consider Figure~\ref{fig:opec_compositional}: panels (i) and (ii) depict training instances with OPEC~2. These are combined by (a) deleting the fact \texttt{maternal\_grandma\_of(ty, joe)} from story~(i), (b) renaming entities in story~(ii)  to align with those in story~(i) (joe1-> joe etc), and (c) adding the story facts from (ii) to (i). Finally renaming all entities, we obtain story~(iii), which is a problem instance with OPEC~4 and is included in the test set. For NoRA v1.1 we did not include any problems with ambiguity. Figure~\ref{fig:nora11} shows the performance of models on NoRA v1.1. The main conclusions from Table \ref{tabMainResults} remain valid. This demonstrates that the inability of models to handle off-path reasoning remains robust to variations in how the problem instances are generated. 

We also introduce another dataset, called \textbf{HetioNet}, which was inspired by \citet{himmelstein2023hetnet}. This dataset is based on a completely different set of world rules, unrelated to family relationships. Here, entities correspond to \emph{diseases}, \emph{genes}, and \emph{drugs}, and relations capture biological phenomena. Moreover, the kinds of regularities that models are expected to learn are rather different. For instance, while families are organized into hierarchical structures, no such structure exists in the case of HetioNet. A detailed analysis of HetioNet is provided in Appendix~\ref{sec:HN}. It shows that, even after this shift in relational regularity, most models continue to struggle on tasks that require off-path reasoning. Surprisingly, however, the EpiGNN outperforms ETs on the test-OPEC split of HetioNet. Further work is needed to better understand the kinds of regularities that different models are able to capture.

\section{Conclusions}
We have introduced a new benchmark for systematic relational reasoning, called NoRA. It has three core features which makes it more challenging than existing benchmarks: the need for off-path reasoning, the presence of ambiguities, and the fact that entities can be simultaneously related in different ways. We found all methods to struggle significantly with off-path reasoning, suggesting that fundamentally different architectures may be needed to push forward the state-of-the-art in neural relational reasoning. Interestingly, Large Reasoning Models such as o3 were not able to solve problem instances that require off-path reasoning either, even when explicitly given all the required rules.
Surprisingly, the presence of ambiguity did not pose any particular challenges for the tested models. However, further analysis revealed this to be due to the presence of shortcuts, allowing models to solve these problem instances without actually needing to reason about ambiguity. This highlights the challenge in generating hard problem instances. Finally, to test the robustness of our findings, we introduced two additional datasets: NoRA v1.1 and HetioNet.

%and can serve as a new gold standard for non-path-like, ambiguous, relational reasoning.
% We also found that the considered reasoning problems are challenging for pretrained Large Reasoning Models such as o3, even when the set of world rules that are needed for reasoning are explicitly given. 
%(in terms of the off-path entity count) proving to be challenging.
% Despite generalizing well with respect to length and therefore the scale of the graph, performance of edge transformers [NOPE... bad taste to call out models by name unless you're openai]
% We also found the need to represent to degenerate edges was crucial for performance when inference hops were high

\paragraph{Limitations}
The ability to measure the difficulty of problem instances is important for testing models in a systematic way. However, metrics such as \emph{reasoning depth} and BL are sensitive to the way in which the knowledge base has been encoded. In the experiments with o3 we  saw examples where the ``true'' reasoning depth was lower than that measured by the metric. All ambiguities are not equally challenging, which is something that \emph{reasoning width} only partially captures. 
% Furthermore, our experiments are limited to a single set of world rules. It would be possible to extend the benchmark with datasets for other sets of world rules. However, while it is straightforward to generate synthetic sets of world rules, designing rules which are both realistic and interesting from a reasoning perspective is considerably harder.

\newpage
\paragraph{Acknowledgements}
 This work was supported by EPSRC grant EP/W003309/1.

% Limitations
%   - Length generalization highly sensitive to logic program encoding
%   - We only test on one set of world rules, but the process we introduced is easily extensible
%    - WE control the marginal in our training sets, to ensure diversity during training. Controlling the joint would be a more genral form systematic reaosning

\bibliographystyle{plainnat}
\bibliography{references}

\newpage
\section*{NeurIPS Paper Checklist}

\begin{enumerate}

\item {\bf Claims}
    \item[] Question: Do the main claims made in the abstract and introduction accurately reflect the paper's contributions and scope?
    \item[] Answer: \answerYes{} % Replace by \answerYes{}, \answerNo{}, or \answerNA{}.
    \item[] Justification: The paper and the abstract are aligned in content and scope. 

\item {\bf Limitations}
    \item[] Question: Does the paper discuss the limitations of the work performed by the authors?
    \item[] Answer: \answerYes{} % Replace by \answerYes{}, \answerNo{}, or \answerNA{}.
    \item[] Justification: There is a paragraph discussing the limitations at the end of the conclusions section.

\item {\bf Theory assumptions and proofs}
    \item[] Question: For each theoretical result, does the paper provide the full set of assumptions and a complete (and correct) proof? \answerNA{}
    \item[] Answer: \answerNA{} % Replace by \answerYes{}, \answerNo{}, or \answerNA{}.
    \item[] Justification: \answerNA{}

    \item {\bf Experimental result reproducibility}
    \item[] Question: Does the paper fully disclose all the information needed to reproduce the main experimental results of the paper to the extent that it affects the main claims and/or conclusions of the paper (regardless of whether the code and data are provided or not)?
    \item[] Answer: \answerYes{} % Replace by \answerYes{}, \answerNo{}, or \answerNA{}.
    \item[] Justification: The details of the dataset generation process, as well as the complete description of our evaluation methodology, are provided in the appendix materials. The training and held-out test datasets are publicly accessible via Hugging Face. As stated during the initial submission, both the dataset URL and the corresponding Croissant metadata file were included. The code used for data generation was also shared at that time and includes straightforward execution instructions. All datasets, along with the code used to generate them, are publicly available. In addition, the code for training and evaluating the models is hosted in a public repository. Links to all these resources are provided directly within the paper.


\item {\bf Open access to data and code}
    \item[] Question: Does the paper provide open access to the data and code, with sufficient instructions to faithfully reproduce the main experimental results, as described in supplemental material?
    \item[] Answer: \answerYes{} % Replace by \answerYes{}, \answerNo{}, or \answerNA{}.
    \item[] Justification: The dataset has been shared. The code used for generating and evaluating the dataset was also shared, with instructions for creating environment and running the python code.

\item {\bf Experimental setting/details}
    \item[] Question: Does the paper specify all the training and test details (e.g., data splits, hyperparameters, how they were chosen, type of optimizer, etc.) necessary to understand the results?
    \item[] Answer: \answerYes{} % Replace by \answerYes{}, \answerNo{}, or \answerNA{}.
    \item[] Justification: Broad details are provided in the main paper. The full details are included in Appendix. Code is available publicly.  %Exact methods to create the confidence interval including the factors of variability that the error bars are capturing. 

\item {\bf Experiment statistical significance}
    \item[] Question: Does the paper report error bars suitably and correctly defined or other appropriate information about the statistical significance of the experiments?
    \item[] Answer: \answerYes{}% Replace by \answerYes{}, \answerNo{}, or \answerNA{}.
    \item[] Justification: This information is in the supplementary materials. 

\item {\bf Experiments compute resources}
    \item[] Question: For each experiment, does the paper provide sufficient information on the computer resources (type of compute workers, memory, time of execution) needed to reproduce the experiments?
    \item[] Answer: \answerYes{} % Replace by \answerYes{}, \answerNo{}, or \answerNA{}.
    \item[] Justification: specifics are included in  Appendix.
    
\item {\bf Code of ethics}
    \item[] Question: Does the research conducted in the paper conform, in every respect, with the NeurIPS Code of Ethics \url{https://neurips.cc/public/EthicsGuidelines}?
    \item[] Answer: \answerYes{} % Replace by \answerYes{}, \answerNo{}, or \answerNA{}.
    \item[] Justification: The research conducted in the paper fully conforms with the Code of Ethics.

\item {\bf Broader impacts}
    \item[] Question: Does the paper discuss both potential positive societal impacts and negative societal impacts of the work performed?
    \item[] Answer: \answerNA{} % Replace by \answerYes{}, \answerNo{}, or \answerNA{}.
    \item[] Justification: We foresee no immediate scope for potential malicious or unintended uses, fairness considerations, privacy considerations, and security considerations.
    
\item {\bf Safeguards}
    \item[] Question: Does the paper describe safeguards that have been put in place for responsible release of data or models that have a high risk for misuse (e.g., pretrained language models, image generators, or scraped datasets)?
    \item[] Answer: \answerNA{} % Replace by \answerYes{}, \answerNo{}, or \answerNA{}.
    \item[] Justification: No pretrained language models, image generators, or scraped datasets are created.

\item {\bf Licenses for existing assets}
    \item[] Question: Are the creators or original owners of assets (e.g., code, data, models), used in the paper, properly credited and are the license and terms of use explicitly mentioned and properly respected?
    \item[] Answer: \answerYes{} % Replace by \answerYes{}, \answerNo{}, or \answerNA{}.
    \item[] Justification:  \textbf{Clingo} (version 5.7.1): We used the Clingo ASP solver~\citep{gebser2011potassco}, available at \url{https://potassco.org/clingo/}. The source code is available at \url{https://github.com/potassco/clingo} and the Python package at \url{https://pypi.org/project/clingo/}. Clingo is distributed under the MIT License.

\item {\bf New assets}
    \item[] Question: Are new assets introduced in the paper well documented and is the documentation provided alongside the assets?
    \item[] Answer: \answerYes{}% Replace by \answerYes{}, \answerNo{}, or \answerNA{}.
    \item[] Justification: The datasets corresponding to the three reasoning worlds---\textsc{NoRA}, \textsc{NoRA-1.1}, and \textsc{InspiredFromHetionet}---are hosted publicly on Hugging Face. Each dataset is accompanied by a validated Croissant metadata file, following the NeurIPS 2025 Datasets and Benchmarks guidelines (\url{https://neurips.cc/Conferences/2025/DatasetsBenchmarks-FAQ}). The three Croissant files were individually validated, packaged into a single ZIP archive, and uploaded as required. A central Hugging Face landing page provides unified access to all three datasets. In addition, the code used to generate the datasets is openly available in a public GitHub repository.
    

\item {\bf Crowdsourcing and research with human subjects}
    \item[] Question: For crowdsourcing experiments and research with human subjects, does the paper include the full text of instructions given to participants and screenshots, if applicable, as well as details about compensation (if any)? 
    \item[] Answer: \answerNA{} % Replace by \answerYes{}, \answerNo{}, or \answerNA{}.
    \item[] Justification:\answerNA{}

\item {\bf Institutional review board (IRB) approvals or equivalent for research with human subjects}
    \item[] Question: Does the paper describe potential risks incurred by study participants, whether such risks were disclosed to the subjects, and whether Institutional Review Board (IRB) approvals (or an equivalent approval/review based on the requirements of your country or institution) were obtained?
    \item[] Answer: \answerNA{} % Replace by \answerYes{}, \answerNo{}, or \answerNA{}.
    \item[] Justification: \answerNA{}

\item {\bf Declaration of LLM usage}
    \item[] Question: Does the paper describe the usage of LLMs if it is an important, original, or non-standard component of the core methods in this research? Note that if the LLM is used only for writing, editing, or formatting purposes and does not impact the core methodology, scientific rigorousness, or originality of the research, declaration is not required.
    %this research? 
    \item[] Answer:  \answerNA{} % Replace by \answerYes{}, \answerNo{}, or \answerNA{}.
    \item[] Justification: \answerNA{}

\end{enumerate}
\newpage
\appendix
\section{Code and Resources}
\label{sec:resources}

The codebase used for generating examples with \textsc{ASP} (Answer Set Programming) is publicly available at:  
\url{https://github.com/axd353/WhenNoPathsLeadToRome.git}. The code for conducting experiments with models such as \textsc{ET}, \textsc{RAT}, and \textsc{EpiGNN}---used to produce the results in Table~\ref{tabMainResults}---is available at:  
\url{https://github.com/erg0dic/whennopathsleadtorome}.

\vspace{0.5em}
The datasets corresponding to the three reasoning worlds ( \textsc{NoRA}, \textsc{NoRA-1.1} and \textsc{InspiredFromHetionet}) can be accessed collectively at:  
\url{https://huggingface.co/datasets/axd353/When-No-Paths-Lead-to-Rome}.

\vspace{0.5em}
For reference, the complete world-rule specifications for each of these worlds are provided at:  
\url{https://github.com/axd353/WhenNoPathsLeadToRome/tree/main/ExplicitWorldRuleFilesForReference}.

\section{Additional experimental results}

\subsection{Main results} 
% (IRTAZA) probably sunset this section
In the main paper, we reported results in terms of exact match (Table~\ref{tabMainResults}). In Table~\ref{tabMainResults-expanded}, we complement this analysis by reporting the results in terms of weighted F1.
The weighted F1-score is calculated as the macro F1-score for each label, aggregated using a weighted mean (based on their frequency in the dataset). Exact-match accuracy requires models to predict all labels correctly when multiple labels are true. The weighted F1 metric still provides positive contribution when at least some labels are predicted correctly accounting for class imbalances. Consequently, this metric can often yield higher scores. This is evident, for instance, in the test-OPEC dataset, where multiple target relations have to be predicted. For example, if the target relations are ``aunt'' and ``maternal aunt'', it may be the case that we only need off-path reasoning for predicting ``maternal aunt''. A model that is not capable of off-path reasoning but that can correctly predict ``aunt'' would thus still be partially rewarded.
%Identifying ``maternal aunt'' requires using off-path information. 
%The results are again in line with the exact match accuracies.

\begin{table}[h]
% \footnotesize
\scriptsize
\centering
\caption{Results of state-of-the-art models for systematic reasoning on the NoRA test sets.\label{tabMainResults-expanded}}
\begin{tabular}{llccccccccc}
\toprule
& & \multicolumn{4}{c}{\textbf{Trained with ambiguity}} & \multicolumn{3}{c}{\textbf{Trained without ambiguity}}\\
\cmidrule(lr){3-6}\cmidrule(lr){7-9}
&  & \textbf{D} & \textbf{W} & \textbf{BL} & \textbf{OPEC} &  \textbf{D-na} & \textbf{BL-na} & \textbf{OPEC-na}\\
\midrule
\multirow{10}{*}{\rotatebox{90}{\textbf{Weighted F1}}}
& ET (single-edge)  & \textbf{0.740} & 0.814 & 0.432 & 0.413  & \textbf{0.816} & \textbf{0.233} & \textbf{0.410} \\
& ET (multi-edge)   & 0.335 & \textbf{0.860} & \textbf{0.888} & \textbf{0.504}  & 0.336 & 0.120 & 0.394 \\
& RAT (single-edge)  & 0.329 & 0.744 & 0.437 & 0.399  & 0.333 & 0.188 & 0.383 \\
& RAT (multi-edge)  & 0.677 & 0.766 & 0.747 & 0.457  & 0.759 & 0.067 & 0.294 \\
& EpiGNN-\texttt{min} (margin)   & 0.326 & 0.296 & 0.082 & 0.116  & 0.326 & 0.112 & 0.206 \\
& EpiGNN-\texttt{min} (BCE)  & 0.625 & 0.633 & 0.316 & 0.180  & 0.319 & 0.049 & 0.218 \\
& EpiGNN-\texttt{mul} (BCE)  & 0.554 & 0.667 & 0.320 & 0.185  & 0.717 & 0.076 & 0.249 \\
 & NBFNet (margin)  & 0.000 & 0.000 & 0.000 & 0.000  & 0.000 & 0.000 & 0.000 \\
 & NBFNet (BCE)  & 0.646 & 0.665 & 0.347 & 0.225  & 0.775 & 0.083 & 0.261 \\
& R-GCN  & 0.691 & 0.455 & 0.189 & 0.286  & 0.704 & 0.122 & 0.195 \\
\bottomrule
\end{tabular}
\end{table}

\subsection{HetioNet}
\label{sec:HN}
To analyze how the results generalize to other datasets, we present results for another world, in addition to NoRA. This world is called Hetionet and is inspired by \citet{himmelstein2023hetnet}.
%to demonstrate the generality of our method for constructing systematic reasoning benchmarks. 

In the HetioNet world, there are three kinds of entities: compounds, diseases, and genes. Compounds and genes can palliate a disease. Compounds can be used to treat a disease, or they can be marked as unusable for treating a disease. Off-path reasoning emerges because, to be used to treat a disease, a compound must both palliate that disease and have no side effects. Compounds and genes can also upregulate a gene; if an entity upregulates a gene that palliates a disease, then the entity itself palliates that disease. Compounds can be similar to each other in three different ways:
\begin{itemize}
    \item  ss2(c1, c2) means that c1 and c2 palliate the same diseases;
    \item ss3(c1, c2) means that c1 and c2 either both have side effects or neither has side effects;
    \item ss1(c1, c2) means that c1 and c2 have the same regulatory properties towards genes.
\end{itemize}

HetioNet has a significantly different regularity than the NoRA world, particularly because there is no hierarchical tree-like structure—two compounds can be related in multiple ways concurrently (whereas in NoRA, your uncle cannot be your brother). We created training sets, test-D, and test-OPEC-na in a similar way to NoRA. As for NoRA v1.1, all examples observed during testing are stitched-up versions of one or more examples that were seen during training. The data are split as follows: OPEC < 3, BL < 1.33, D < 6 for the train splits and OPEC = 3 for Test-OPEC, D = 7 for Test-D. There are no problem instances with ambiguitiy in the case of HetioNet.

The HetioNet world contains far fewer rules (55) compared to NoRA (284). Moreover, only two to three types of relations can exist between entities of two types—for example, a compound may either upregulate or downregulate a gene. In contrast, numerous types of relationships can hold between two entities that are persons in NoRA. Consequently, HetioNet represents a much simpler and more easily solvable world for most models.

The results for state-of-the-art models are shown in Table~\ref{tab:hetionet-results}. In line with the results for NoRA, the edge transformer emerges as the best performing model for the in-distribution set set and the Test-D test set. However, the EpiGNN-\texttt{min} model has the best performance on Test-OPEC, presumably due to the strong inductive bias of the min pooling operator in this world.

\begin{table}[h]
\footnotesize
% \scriptsize
\centering
\caption{Results of state-of-the-art models on the HetioNet test sets.}
\label{tab:hetionet-results}
\begin{tabular}{lcccccc}
\toprule
 & \multicolumn{3}{c}{\textbf{Accuracy (Exact Match) }} & \multicolumn{3}{c}{\textbf{Weighted F1}}\\
\cmidrule(lr){2-4}\cmidrule(lr){5-7}
 & \textbf{In dist.} & \textbf{D} & \textbf{OPEC} & \textbf{In dist.} & \textbf{D} & \textbf{OPEC}\\
\midrule
ET (single-edge)     & \textbf{0.838} & \textbf{0.907} & 0.495 & \textbf{0.936} & \textbf{0.958} & 0.721 \\
ET (multi-edge)    & 0.725 & 0.845 & 0.486 & 0.857 & 0.887 & 0.680 \\
RAT (single-edge)       & 0.657 & 0.756 & 0.466 & 0.831 & 0.811 & 0.687 \\
RAT (multi-edge)      & 0.784 & 0.671 & 0.641 & 0.908 & 0.785 & 0.768 \\
R-GCN                  & 0.712 & 0.356 & 0.541 & 0.901 & 0.500 & 0.847 \\
NBFNet (BCE)                & 0.742 & 0.428 & 0.576 & 0.879 & 0.571 & 0.757 \\
EpiGNN-\texttt{min} (BCE)      & 0.714 & 0.351 & \textbf{0.772} & 0.837 & 0.365 & \textbf{0.863} \\
EpiGNN-\texttt{mul} (BCE)      & 0.704 & 0.499 & 0.624 & 0.832 & 0.615 & 0.812 \\
\bottomrule
\end{tabular}
\end{table}

\subsection{In-depth analyses for other baselines}
\begin{figure}
    \centering
    \includegraphics[width=1\linewidth]{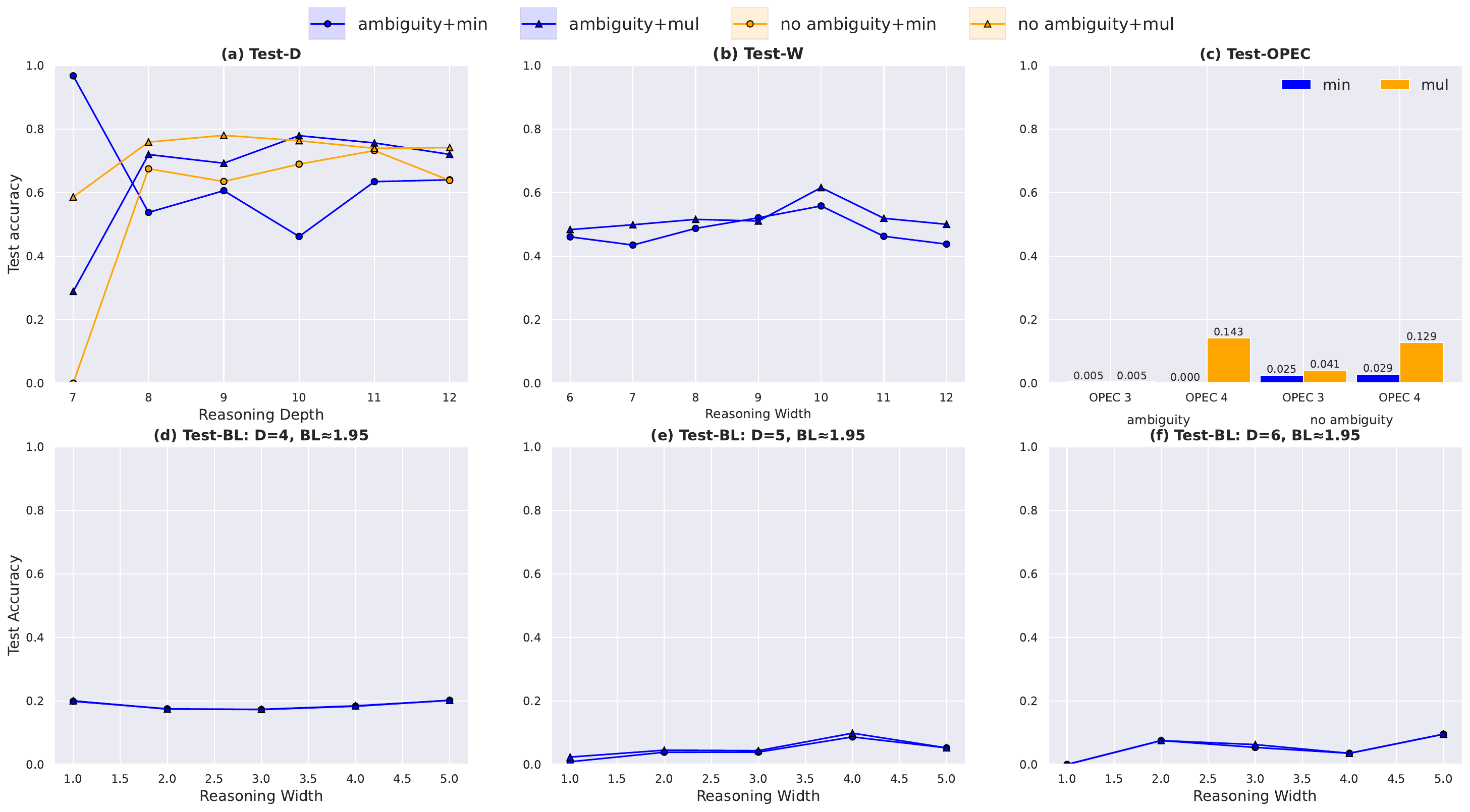}
    \caption{Analysis of the performance of EpiGNN on various splits of the dataset. Weighted F1 scores per class are computed to avoid class imbalances affecting the metric score for the various test splits. }
    \label{fig:depth-first-epignn-2}
\end{figure}

\begin{figure}
    \centering
    \includegraphics[width=1\linewidth]{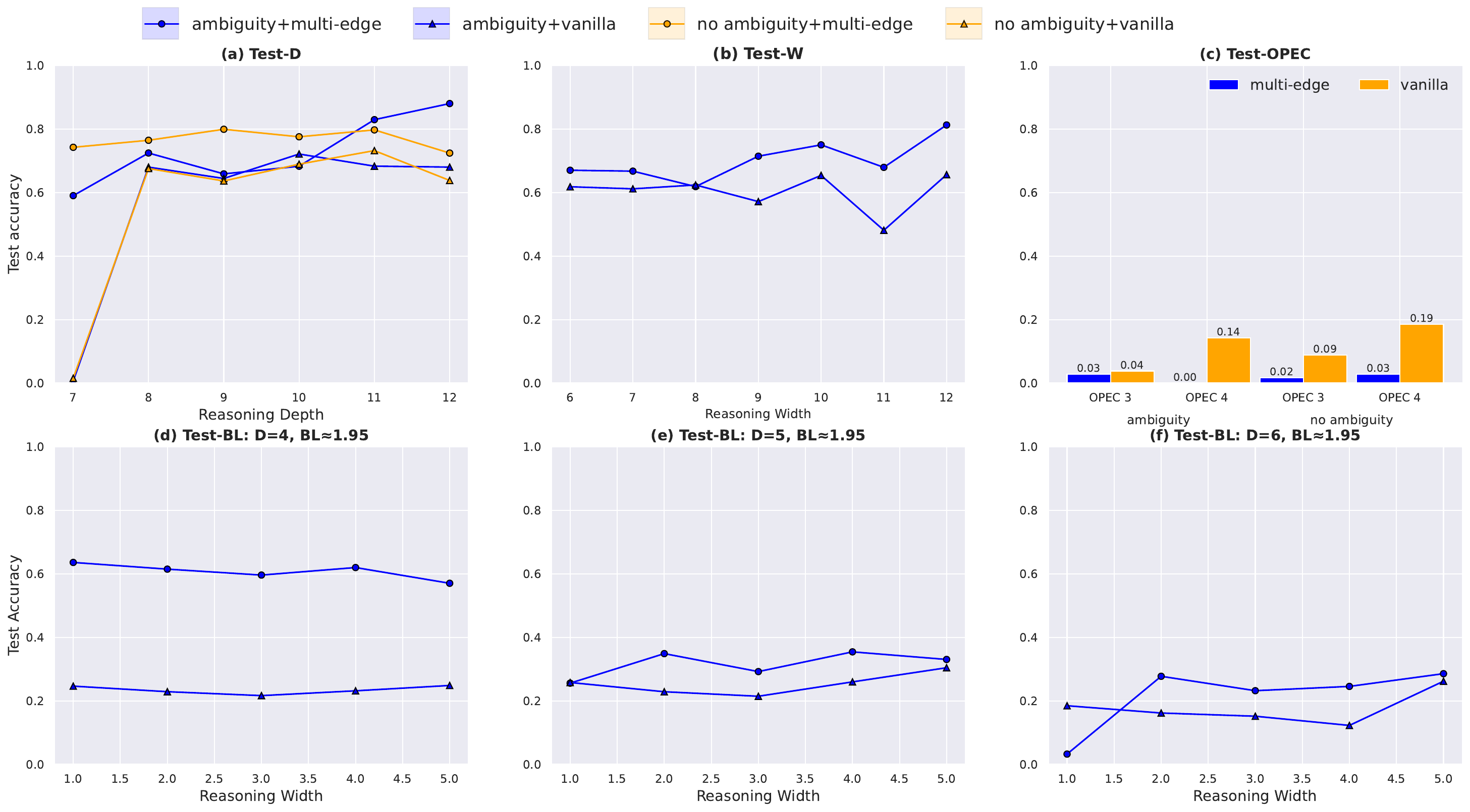}
    \caption{Analysis of the performance of RAT on various splits of the dataset. Weighted F1 scores per class are computed to avoid class imbalances affecting the metric score for the various test splits. }
    \label{fig:depth-first-rat-2}
\end{figure}

We provide further analysis for a GNN (EpiGNN) in Figure~\ref{fig:depth-first-epignn-2} and for RAT in Figure~\ref{fig:depth-first-rat-2}, to complement the analysis of edge transformers in the main text. Broadly, the trends observed in the main text hold for other models with respect to length and width generalization. For RAT, the single-edge or vanilla model has a higher OPEC performance than it multi-edge counterpart in Figure~\ref{fig:depth-first-rat-2}(c). Also, the multi-edge RAT is better at width generalization in figures~\ref{fig:depth-first-rat-2}(d)-(f). We also show a weighted F1 that overcomes class imbalances for some figures which highlights a similar trend to the accuracy curves. For the EpiGNN, the \texttt{mul} aggregation function does notably better than \texttt{min} for OPEC in figures~\ref{fig:depth-first-epignn-2}(c) and also on the Test-D split in figures~\ref{fig:depth-first-epignn-2}(a). 

% \begin{figure}
%     \centering
%     \includegraphics[width=1\linewidth]{figs/edge_t_depth_first_3x3.pdf}
%     \caption{Depth-first analysis of the performance of ETs on various splits of the dataset. Weighted F1 scores per class are computed to avoid class imbalances affecting the metric score for the various test splits. }
%     \label{fig:depth-first-et-new-2}
% \end{figure}

\section{Notation and task: Intuitive walkthrough}
\label{seec:IntuitiveWalkthrough}
Here we give an intuitive overview using examples instead of formal definitions for the notations introduced formally in the main paper. We focus on stories without ambiguity, as ambiguity is discussed in detail in Appendix \ref{sec:AmbFacts}.
We use \textbf{Answer Set Programming (ASP)} as the underlying language to encode problem instances in NoRA. %We also borrow from ASP syntax \cite{lifschitz2019answer, gebser2011potassco} to describe the dataset structure.

The dataset is composed of three parts: \textbf{world rules}, \textbf{stories}, and \textbf{entailed atoms}. The \emph{world rules} define the underlying regularity of relationships in a given universe. These rules are not exposed to the model. The goal of learning models is to infer these hidden rules through example instances and apply them to reasoning tasks. We consider two sets of world rules (i.e.\ two worlds):
\begin{description}
\item[NoRA-mini:] A simplified world used for illustrative purposes. 
\item[NoRA-full:] A richer and more fine-grained world with a broader set of rules, used to generate the full benchmark.
\end{description}

\paragraph{World rules} Figure~\ref{fig:kanr_fig_full}(a) shows an example of the world rules in NoRA-mini. These rules fall into three categories: definite rules, constraints and facts. 

A \textbf{definite rule} consists of a \textbf{body} and a \textbf{head}. The body is a conjunction of one or more atoms; the head is a single atom. In the absence of constraints, we can think of these rules in terms of standard implication: if all atoms in the body are true, then the head must also be true. For example, consider the following rule from Figure~\ref{fig:kanr_fig_full}(a):

\begin{align*}
	\texttt{living\_in\_same\_place(X,Z) :-} &\texttt{living\_in\_same\_place(X,Y),} \\
    &\quad \texttt{living\_in\_same\_place(Y,Z).}
\end{align*}

This rule states: for any entities \texttt{X}, \texttt{Y}, and \texttt{Z}, if \texttt{X} lives in the same place as \texttt{Y}, and \texttt{Y} lives in the same place as \texttt{Z}, then \texttt{X} also lives in the same place as \texttt{Z}.  

\textbf{Constraints} are rules without a head. They specify sets of atoms that are \emph{not} allowed to be simultaneously true. For example:

\begin{quote}
	\texttt{:- belongs\_to(X, underage), parent\_of(X, Y).}
\end{quote}

This constraint expresses that an underage person cannot be a parent. Note, in our notation, \texttt{rel(X, Y)} means \texttt{X} is \texttt{rel} of \texttt{Y}. So, \texttt{parent\_of(X, Y)} means X is the parent of Y. 

\textbf{Facts} are atoms that are always true. They are rules without a body and are often used to declare properties of constants. For example:

\begin{quote}
	\texttt{is\_agegroup(underage).}
\end{quote}

\paragraph{Stories} Each story consists of a set of \emph{story facts}, which are grounded atoms, i.e., they contain no variables. For example, in Figure~\ref{fig:kanr_fig_full}(b), the fact:

\begin{quote}
	\texttt{school\_mates\_with(ram, irfan).}
\end{quote}

states that \texttt{ram} and \texttt{irfan} are schoolmates. Combining the story facts with the fixed world rules one obtains a logic program. Abusing terminology, we sometimes call this logic program the story. 

\paragraph{Entailed atoms via stable models}  Stable models/answer sets are the solution of a logic program. Intuitively, they are a minimal set of atoms/facts that follow from the logic program (see Section \ref{sec:stab_mod} for formal definitions). A stable model includes both the explicitly stated story facts and additional possible atoms that follow logically. These additional atoms are called \textbf{entailed atoms}. Figure~\ref{fig:kanr_fig_full}(c) shows the stable model of the story from Figure~\ref{fig:kanr_fig_full}(b). The entailed atoms are highlighted. If an entailed atom has a binary predicate (relationship), its first argument is called the source entity and its second argument the target entity.

\paragraph{Example format and reasoning task} While the world rules are kept fixed, multiple logic programs are generated by randomly sampling many sets of story facts. For each such program, the corresponding entailed atoms are computed.

An individual \textbf{example} in the dataset consists of:
\begin{itemize}
	\item The story facts (input), encoded as a graph.
	\item The target and source entities of an entailed atom, which define the query. % where the predicate of this entailed atom is to be inferred.\todo{This is only true for the single-label case, and so doesn't agree with what we wrote in the paper. We should either update this, or remove it (since these details are in the paper anyway).}
\end{itemize}
Let $a$ and $b$ be the atoms defined in the query. 
The task is to predict all relations $r$ such that $r(a,b)$ can be entailed from the story facts. 
%It is possible for multiple relationships to be True, in this case the task is to find all possible predicates that capture relationships between the source and target entities. 
For the example in Figure~\ref{fig:kanr_fig_full}(d), the entailed atom is \texttt{living\_in\_same\_place(irfan, lola)}. A model attempting to solve NoRA will be shown the story-facts, the source entity \texttt{irfan}, and the target entity \texttt{lola}, and it must infer all predicates/relationships   (including missing ones  which is only \texttt{living\_in\_same\_place} in this case) between source and target . In NoRA-full, multiple relationships/predicates might be true between the same two entities.

\paragraph{Reasoning depth} The difficulty of deriving an entailed atom is influenced by the number of reasoning steps required to reach it, excluding the direct use of story-facts. For example, Figure~\ref{fig:kanr_fig_full}(e) shows that for the given story in NoRA-mini, deriving \texttt{living\_in\_same\_place(irfan, lola)} requires six inference steps. Since derivations may not be unique, we use derivations that are minimal (in a sense) to calculate the metric called \textbf{reasoning depth}.

\begin{figure}[t]
	\centering
	
	% Left column: (a)
	\begin{minipage}[t]{0.48\textwidth}
		\footnotesize
		\textbf{(a) World Rules} \\[0.1em]
		\textbf{Definite Rules} \\
		\textcolor{violet}{
		\texttt{living\_in\_same\_place(Y, X) :- school\_mates\_with(Y, X).} \\
		\texttt{living\_in\_same\_place(Y, X) :- belongs\_to(X, underage), parent\_of(Y, X).} \\
		\texttt{living\_in\_same\_place(Y, X) :- living\_in\_same\_place(X, Y).} \\
		\texttt{living\_in(Y, Z) :- living\_in\_same\_place(X, Y), living\_in(X, Z).} \\
		\texttt{belongs\_to(X, underage) :- school\_mates\_with(X, U).} \\
		\texttt{living\_in\_same\_place(X,Z) :- living\_in\_same\_place(X,Y), living\_in\_same\_place(Y,Z).}} \\[0.1em]
		\textbf{Constraint} \\
		{\texttt{:- belongs\_to(X, underage), parent\_of(X, Y).}}\\[0.1em]
		\textbf{Facts} \\
		\textcolor{green!50!black}{\texttt{is\_agegroup(underage).}} 
	\end{minipage}
	\hfill
	% Right column: (b) + (c)
	\begin{minipage}[t]{0.48\textwidth}
		\footnotesize
		\textbf{(b) Story Facts} \\[0.3em]
		\textcolor{green!50!black}{\texttt{school\_mates\_with(ram, irfan).}} \\
		\textcolor{green!50!black}{\texttt{parent\_of(lola, ram).}} \\
		\textcolor{green!50!black}{\texttt{living\_in(irfan, calcutta).}}
		
		\vspace{1em}
		\textbf{(c) Stable Model} \\
		\quad \textcolor{red}{\texttt{living\_in\_same\_place(ram, irfan),}} \\
		\quad \textcolor{red}{\texttt{living\_in\_same\_place(lola, ram),}} \\
		\quad \textcolor{red}{\texttt{living\_in\_same\_place(ram, lola),}} \\
		\quad \textcolor{red}{\texttt{living\_in\_same\_place(irfan, ram),}} \\
		\quad \textcolor{red}{\texttt{living\_in\_same\_place(lola, irfan),}} \\
		\quad \textcolor{red}{\texttt{living\_in\_same\_place(irfan, lola),}} \\
		\quad \textcolor{red}{\texttt{living\_in\_same\_place(ram, ram),}} \\
		\quad \textcolor{red}{\texttt{living\_in\_same\_place(lola, lola),}} \\
		\quad \textcolor{red}{\texttt{living\_in\_same\_place(irfan, irfan),}} \\
		\quad \texttt{school\_mates\_with(ram, irfan),} \\
		\quad \texttt{parent\_of(lola, ram),} \\
		\quad \textcolor{red}{\texttt{belongs\_to(ram, underage),}} \\
		\quad \texttt{living\_in(irfan, calcutta),} \\
		\quad \textcolor{red}{\texttt{living\_in(ram, calcutta),}} \\
		\quad \textcolor{red}{\texttt{living\_in(lola, calcutta),}} \\
		\quad \texttt{is\_agegroup(underage)}. 
	\end{minipage}
	
	\vspace{1.5em}
	
	% Graph: (d)
	\begin{minipage}[t]{0.9\textwidth}
		\centering
		\footnotesize
		\textbf{(d) Visualizing the  Reasoning Task} \\
		\begin{tikzpicture}[node distance=2.5cm, every node/.style={font=\small}]
			\node[draw, circle] (lola) {Lola};
			\node[draw, circle, right of=lola] (ram) {Ram};
			\node[draw, circle, right of=ram] (irfan) {Irfan};
			\node[draw, rectangle, right of=irfan, xshift=2.5cm] (calc) {Calcutta};
			
			\draw[->, thick] (lola) to[bend left] node[above] {\texttt{parent\_of}} (ram);
			\draw[->, thick] (ram) to[bend left] node[above] {\texttt{school\_mates\_with}} (irfan);
			\draw[->, thick, dashed, red] (lola) to[bend right=20] node[below] {\texttt{living\_in\_same\_place}} (irfan);
			\draw[->, thick] (irfan) -- (calc) node[midway, above] {\texttt{living\_in}};
		\end{tikzpicture}
	\end{minipage}
	\vspace{1.5em}	
	% Reasoning Derivation: (e)
\begin{minipage}[t]{0.9\textwidth}
	\centering
	\footnotesize
	\textbf{(e) Derivation for \texttt{living\_in\_same\_place(irfan, lola)}} \\[0.5em]
	\begin{flushleft}
		\textcolor{gray}{\texttt{Fact: school\_mates\_with(ram, irfan)}} \\
		\texttt{1. living\_in\_same\_place(ram, irfan) :- school\_mates\_with(ram, irfan).} \\[0.5em]
		
		\texttt{2. living\_in\_same\_place(irfan, ram) :- living\_in\_same\_place(ram, irfan).} \\[0.5em]
		
		\texttt{3. belongs\_to(ram, underage) :- school\_mates\_with(ram, irfan).} \\[0.5em]
		
		\textcolor{gray}{\texttt{Fact: parent\_of(lola, ram)}} \\
		\texttt{4. living\_in\_same\_place(lola, ram) :- belongs\_to(ram, underage),} \\
		\hspace*{3em} \texttt{parent\_of(lola, ram).} \\[0.5em]
		
		\texttt{5. living\_in\_same\_place(ram, lola) :- living\_in\_same\_place(lola, ram).} \\[0.5em]
		
		\texttt{6. living\_in\_same\_place(irfan, lola) :- living\_in\_same\_place(irfan, ram),} \\
		\hspace*{3em} \texttt{living\_in\_same\_place(ram, lola).}
	\end{flushleft}
\end{minipage}

	\caption{Illustration of (a) world rules, (b) story facts, (c) stable model with entailed atoms in \textcolor{red}{red}, and (d) visual reasoning task: \textit{What is the relationship between Lola and Irfan?} Correct answer: \texttt{living\_in\_same\_place}. (e) Reasoning depth for the entailed atom in d.}
	\label{fig:kanr_fig_full}
\end{figure}

\section{Data generation and sampling}  
\label{sec:data_generation}  

\paragraph{Data generation process}  
We generate approximately 500{,}000 example instances by repeatedly sampling random story facts, as detailed below. The same set of world rules is used for all stories (see \url{https://huggingface.co/datasets/axd353/When-No-Paths-Lead-to-Rome} for the full set of NoRA rules). A single logic program may have multiple entailed atoms, and hence gives rise to multiple example instances in the final dataset. Each story contains two types of entities: \textit{persons} and \textit{places}. First, all entities are generated and assigned a type (person or place). This assignment is governed by a parameter called \texttt{person\_percent}, which determines the probability that an entity is a person. Higher values of \texttt{person\_percent} result in more persons, while lower values yield more places. The value of \texttt{person\_percent} for each story is recorded and included in the dataset.  When a person entity is introduced, its gender is either assigned (male or female) or left unspecified. This is controlled by a per-story parameter called \texttt{no\_gender\_assign}, which captures the proportion of person entities with unspecified gender.

Relationships are sampled from the list of binary predicates defined in the world rules . For each story fact, a predicate is sampled and applied to a randomly chosen pair of entities, resulting in a fact of the form \texttt{rel(e1,e2)} being added to the story. After each fact is added, the resulting logic program is solved to ensure that at least one answer set exists—i.e., that the story remains consistent. If the added fact introduces a contradiction, it is discarded and a new one is sampled instead. The number of entities per story is sampled uniformly between 20 and 50, and the total number of story facts per instance ranges from 30 to 75. Details of how ambiguous facts are introduced into the stories are provided in Section~\ref{sec:AmbFacts}.

While there are many possible relationships that can hold between two people, we only consider one relationship between people and places, namely  \texttt{living\_in}. We thus need to make sure that queries where the source entity is a person and the target entity is a place are not trivial to answer, i.e.\ that models cannot rely on the shortcut that in such cases the answer is always the singleton $\{\texttt{living\_in}\}$. To this end, we have introduced an additional predicate \texttt{not\_living\_in}, which is inferred by the following rule, encoding the fact that a person can only live in one place:
\begin{align*}
\texttt{not\_living\_in(X,Z) :-  living\_in(X,Y), Y $\neq$ Z}
\end{align*}
%If \enquote{a} is a \textbf{person} entity, \enquote{b} is a \textbf{place} entity, and \enquote{c} is another \textbf{place} entity that appears in some story fact and is not equal to \enquote{b}, and \texttt{living\_in(a,b)} is an entailed atom, we add \texttt{not\_living\_in(a,c)} as a query with this story. This is so models training on NoRA cannot use the shortcut that whenever \enquote{a} is a person and \enquote{b} is a place, the only predicate that completes \enquote{a}--\enquote{b} in test and train examples is \texttt{living\_in}. 

\paragraph{Dataset construction}  
From this pool of example instances, we construct training and testing datasets under the constraint that \textit{all target query relationships in the test sets must appear in the training data}. To balance the distribution of problem difficulty in the training set, we use \textit{inverse transform sampling}. A general discussion of the nuances of re-sampling techniques can be found in \citep{levina2017subsampling,das2022monte}. We use rejections sampling, enabling stratified sampling via quantile functions to obtain the training set. See discussion below:

\paragraph{Difficulty stratification}  
Examples are binned by four metrics:  
\begin{itemize}  
  \item \textbf{Reasoning Depth} (3 bins uniformly covering the range), 
  \item \textbf{Reasoning Width} (3 bins uniformly covering the range), 
  \item \textbf{Branching Length (BL)} (2 bins uniformly covering the range),  
  \item \textbf{OPEC} (2 bins: 0 vs. 1--2).  
\end{itemize}  
The sampling process follows a rejection-based strategy, beginning with a large pool of candidate examples and iteratively removing samples to achieve a balanced marginal distribution over difficulty metrics. We aim to balance the dataset along several predefined difficulty axes, denoted as $S_{\text{diff}}$. Each axis in $S_{\text{diff}}$ corresponds to a difficulty metric—such as reasoning depth, branching length (BL), or OPEC—and is associated with a specific number of target bins.

Sampling proceeds in multiple passes (up to a maximum of \texttt{max\_p}), terminating early if a satisfactory balance is achieved. In each pass, the following steps are performed:

\begin{enumerate}
  \item \textbf{Score Initialization:} Initialize a removal score of zero for all examples.
  
  \item \textbf{Metric-wise Imbalance Scoring:} For each difficulty metric $p_{\text{dm}} \in S_{\text{diff}}$, bin the dataset into \texttt{num\_bins[$p_{\text{dm}}$]} quantile-based bins. Identify the over-represented bins (i.e., bins whose sample count exceeds the target). For every example in an over-represented bin, increment its score by one.
  
  \item \textbf{Overrepresentation Removal:} After processing all metrics, make a second pass over $S_{\text{diff}}$. For each over-represented bin, identify examples with the highest accumulated scores and remove them first.
\end{enumerate}

\begin{figure}[h]  
  \centering  
  \begin{subfigure}{0.9\textwidth}  
    \includegraphics[width=\linewidth]{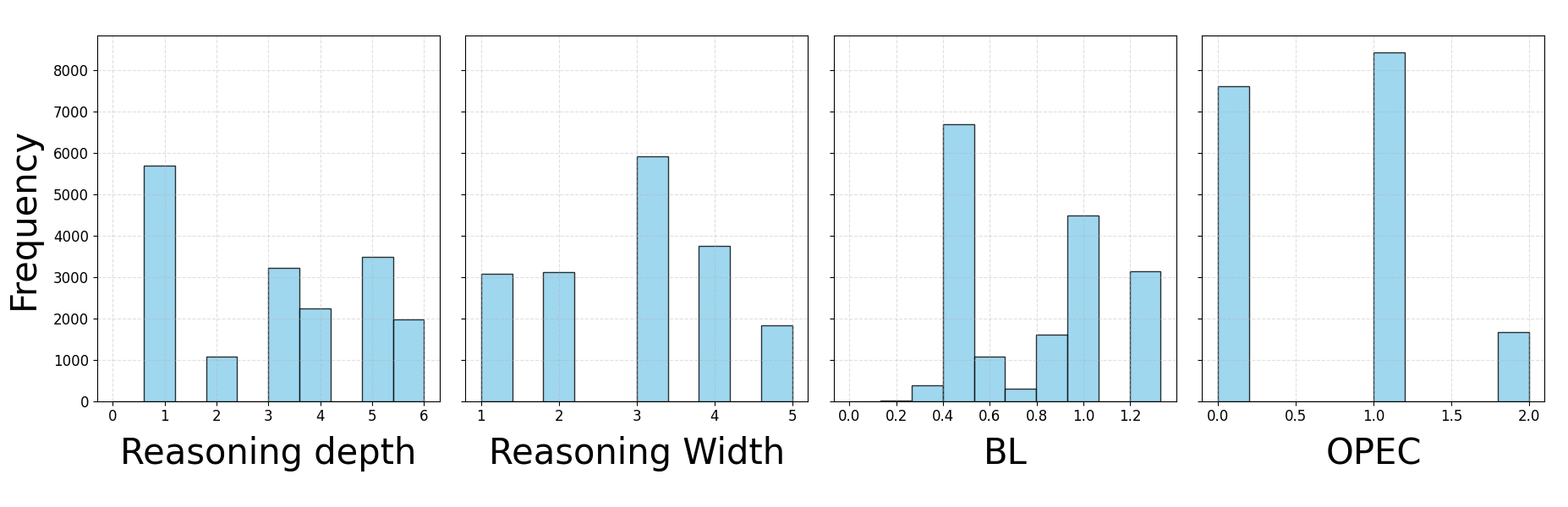}  
    \caption{Train-A (with ambiguity)}  
    \label{fig:train_a}  
  \end{subfigure}  
  \hfill  
  \begin{subfigure}{0.9\textwidth}  
    \includegraphics[width=\linewidth]{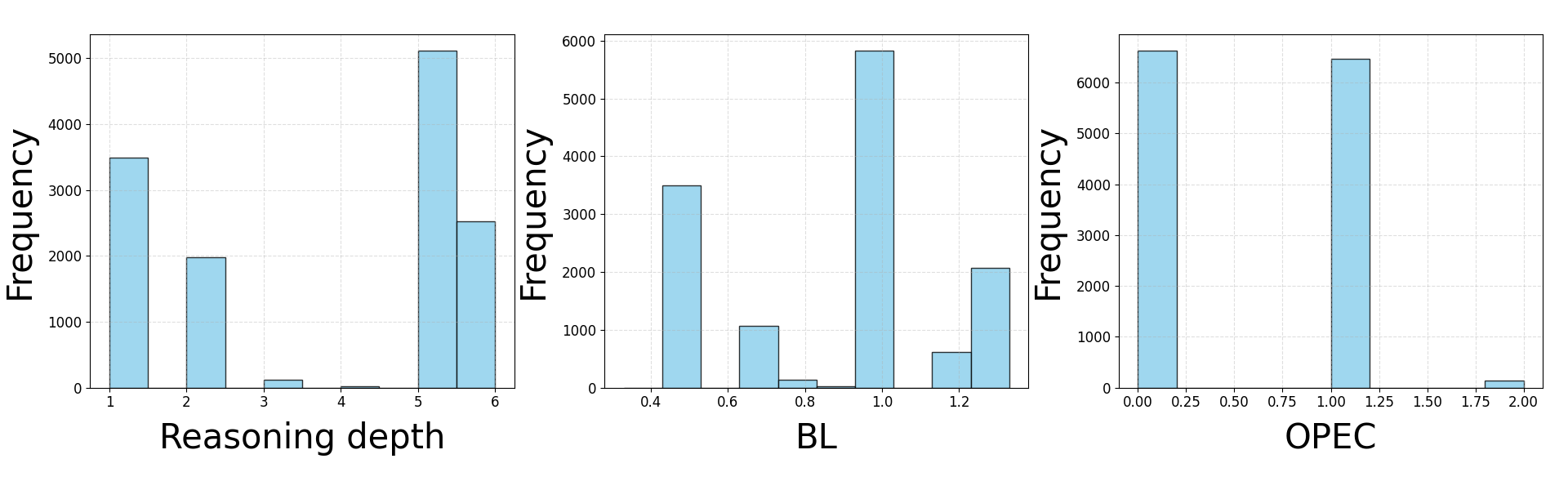}  
    \caption{Train-NA (no ambiguity)}  
    \label{fig:train_na}  
  \end{subfigure}  
  \caption{Distributions of difficulty metrics across training sets. }  
  \label{fig:training_distributions}  
\end{figure}

\paragraph{Training Data distributions}  
This two-pass process, repeated across sampling rounds, ensures that examples contributing disproportionately to skewed distributions are pruned while maintaining as much diversity and coverage as possible. 
Figures~\ref{fig:training_distributions} show the difficulty metric distributions for Train-A (ambiguous) and Train-NA (non-ambiguous) sets.  

\paragraph{Held-out test data}
As mentioned in the main text, we evaluate on various held-out test datasets, where each test dataset is designed to be hard according to one difficulty metric while remaining in-distribution (compared to the training set) in terms of other difficulty metrics. 
For the test datasets with ambiguity:
\begin{itemize}
    \item For \texttt{Test-D}, we ensure that a positive refinement (refinements that have an answer set) has a reasoning depth greater than 6. This is to ensure the problem is actually difficult, as models often take shortcuts by ignoring the derivation of the contradiction in other refinements.    
    \item Likewise, for \texttt{Test-BL} and \texttt{Test-OPEC}, we make sure a positive refinement has $\text{BL} > 1.5$ and $\text{OPEC} \geq 3$, respectively.
\end{itemize}

\section{NoRA rules reflect real-world intuitions}
\label{sec:NoRARealWorldRuleBase}  
We contend that the world rules from
%that are to be learned by systematic reasoning models attempting 
NoRA are \emph{realistic}, in that human beings are able to intuitively accept them to be true (or at least plausible). 
We believe this is a useful feature of our dataset, as it makes it easier to compare neural reasoning models, such as the ones we discuss in this paper, with LLM based approaches.
To test our hypothesis that the rules are realistic, we used an LLM, namely o4-mini, to complete the 284 rules with zero-shot prompting in an open-ended question answering format. Specifically, given the body of a rule, we asked the model the predict the head.

Here are the results for the three types of NoRA rules:
\begin{itemize}
    \item Rule type: \textbf{constraint} \hfill 89.0/90.0 (98.9\%) correct
    \item Rule type: \textbf{definite\_rule} \hfill 184.0/194.0 (94.8\%) correct
    \item  Overall accuracy: 96.1\%
\end{itemize}

%\subsection{Prompts Used}
The prompt we used first defines all predicates:

\begin{verbatim}
Here are some Predicate Definitions:
- "child_of(X,Y)": "X is a child of Y. Order matters: the first argument is 
  the child, the second is the parent"
...
[all predicates are likewise described]
\end{verbatim}

The next part of the prompt differs for rules and constraints.
%
%\subsubsection{Definite Rules Prompt}
For definite rules, where the head is a binary predicate, we use the following prompt:
\begin{verbatim}
Given that all of the following atoms are true:
grandparent_of(X,Y), belongs_to_group(X, male)

What is the relationship between X, Y?
Provide only the predicate with variables in exactly this format:
rel(X,Y)
What is the predicate name that should replace `rel'? Your response should
be rel(X,Y), where rel is your guess. If you think multiple predicates
could work, you must choose the most specific one. For example:
- If both brother and sibling are suitable, choose brother as it's more
specific.
\end{verbatim}

%\subsubsection{Constraint Prompt}
\smallskip
For a constraint, we instead use the following prompt:
\begin{verbatim}
Given that all of the following atoms are true:
has_property(Y, no_daughters), daughter_of(X,Y)

Can this combination of facts logically exist?
Answer exactly one of:
[Possible] [Impossible] [Inevitable]
\end{verbatim}

As a sanity check, we also tested the LRM (o4-mini) with 90 random constraints not in the NoRA rules, but which use the same predicates. Each of these non-world constraints is a slight modification of NoRA constraints. For example, while \enquote{\texttt{:- aunt\_or\_uncle\_of(Y,X), grandchild\_of(Y,X).}} is a NoRA constraint stating someone's aunt cannot also be their grandchild, we modified it to the non-world constraint \enquote{\texttt{:- aunt\_or\_uncle\_of(Y,X), grandchild\_of(U,V)}.} This should be possible as U,V and X,Y can be different pairs of people, and thus in the real world there is no obstruction for this to be true. We note the o4-model's response with the same constraint prompts as above for these non-world constraints: the model \textbf{always responded with ``[Possible]''.}

\section{Experiments with trainable relational reasoning models}
\subsection{Loss functions}
\paragraph{Margin loss}
Let us write $\mathbf{x_{i}}$ to denote the prediction that is obtained by the model for training example $i$, and let $\mathbf{r_{i}}$ denote the embedding of relation $r_i$. We write $\mathbf{r_{i}'}$ to denote some negative example, i.e.\ $\mathbf{r_{i}'}$ for some $r_i'\in\mathcal{R}\setminus\{r_i\}$, the set of all possible relations in NoRA. In the case of multiple target relation vectors, $\mathbf{r_{i}}$, we take the average, $\mathbf{\bar{r}_{i}}$. The overall loss function is:
\begin{align}\label{eqLossmargin}
\mathcal{L}_{\text{margin}} = \sum_{i \in \mathcal{D}}  \max\Big(0, \text{CE}(\mathbf{x_{i}},\mathbf{\bar{r}_{i}}) - \text{CE}(\mathbf{x_{i}},\mathbf{r_{i}'}) + \Delta\Big)
\end{align}
where CE is the cross entropy function and $\Delta$ is the margin value that is set to 1.0 after hyperparameter tuning. The margin loss over multiple models involves an additional sum over the cross entropy differences predicted target relation per model inside the max. At inference time, the target relation is predicted using the negative cross entropy as a score function, with respect to every relation vector in $\mathcal{R}$.

\paragraph{Multi-label binary cross entropy}
We use a multi-label version of the Binary Cross Entropy (BCE) loss for the multi-label classification setting for all NoRA problems. The logits for each class are transformed using a sigmoid function and then the problem is treated as a binary classification problem with a multi-hot target binary vector.
\begin{align}\label{eqbce}
\mathcal{L}_{\text{BCE}} = \sum_{i \in \mathcal{D},j \in \mathcal{R}} \text{CE}(\sigma(x_{ij}), y_{ij})
\end{align}
where $i$ is the sample index, $j$ is the relation index, $x_{ij}$ is the predicted logit and the $y_{ij}$ is the one-hot target class label.

\subsection{Initialization and compute}
All trainable parameters for the models are uniformly initialized. All baseline results that were obtained by us were hyperparameter-tuned using grid search, as detailed below. 
All experiments were conducted using RTX 4090 GPUs. A single experiment using the trainable models can be conducted within a few minutes to 1 hour on a single GPU. This includes training and testing a single model on any test split of NoRA. A single hyperparameter set evaluation is done on about 20\% of the total epochs and training data compared to a full experiment and would take a commensurate amount of time.

\subsection{Hyperparameter settings}
We use the Adam optimizer~\citep{kingma2017adam}. All the models were hyperparameter tuned using an economical grid search over key parameters. For ET and RAT, a grid search was performed over the number of attention heads, hidden dimension size, the number of message passing rounds, and dropout rate. For the GNNs, we grid searched over the hidden dimension size, the number of message passing rounds. In addition for EpiGNN, we also tuned the number of facets. All the optimal hyperparameters are available in the companion code with the manuscript.    
% The number of layers of the \steven{EpiGNN} model is fixed to 9 and the number of negative examples per instance is fixed as 1. The other hyperparameters of the \steven{EpiGNN} model are tuned using grid search. 
% The optimal values that were obtained are mentioned in Table~\ref{hyperparameter}.
% The optimal values that were obtained are mentioned in Table~\ref{hyperparameter}.

\section{Experiments with large reasoning models}
\label{sec:lrmexp}
\begin{figure}[t]
    \centering
    \includegraphics[width=0.7\linewidth]{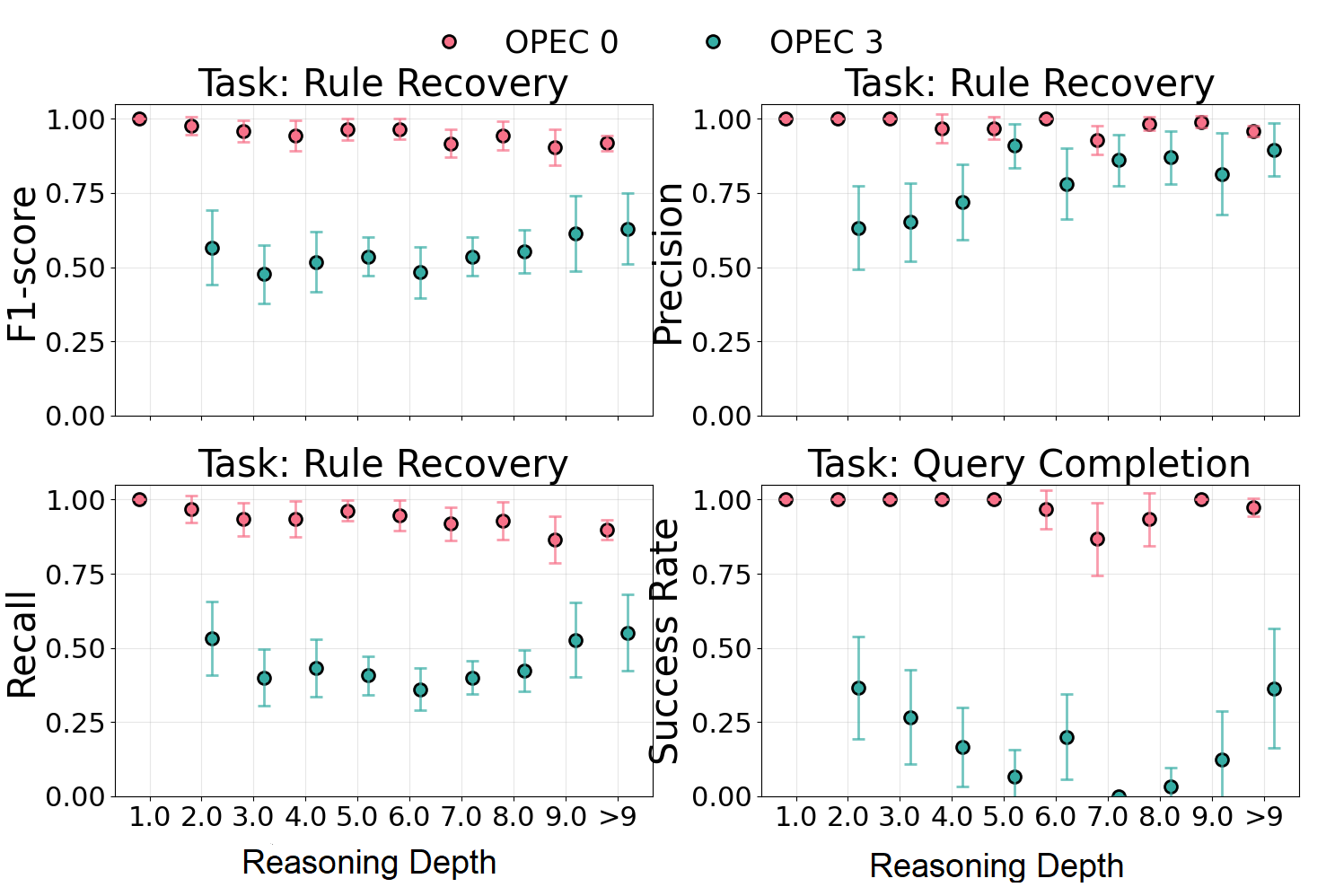}
    \caption{Performance of OpenAI's o3 model on Query Completion and Rule Recovery Tasks. Results separated according to OPEC and the reasoning depth of examples. }
    \label{fig:rule_recovery}
\end{figure}
\paragraph{Rule recovery task}
In addition to the results presented in Section 5 of the main paper on the performance of Large Reasoning Models (LRMs), we evaluate LRM models on a second diagnostic task. Since all NoRA world rules are provided to the model, we additionally task the LRM with outputting the complete set of world rules it used to solve the given query completion task. We call this the \textit{Rule Recovery Task}. Successful completion of the query completion task \textit{without} correct rule recovery indicates that the model may be taking shortcuts to arrive at the correct answer without following the intended reasoning steps.

Figure~\ref{fig:rule_recovery} presents our results side-by-side with the query completion results (a copy of  Figure \ref{fig:opec_query} from the main paper), for easy comparison. This parallel presentation is particularly informative as both tasks are evaluated on identical example instances. The results reveal that while models may have good precision on the rule recovery task, recalling all applicable rules proves substantially more difficult, especially in cases requiring reasoning with significant off-path complexity. The models are evaluated on examples such as those in Figure \ref{fig:opec_query}  of the main paper. 

For Figure \ref{fig:opec_query} in the main paper, the mean success rate and its 95\% confidence interval are estimated using bootstrapping. For Figure \ref{fig:opec_comparisons}, the performance of the o3 variant is assessed across different reasoning depths. Mean success rates are computed as sample averages, with confidence intervals derived via normal approximation using standard deviation estimates from a Binomial parameterization.

\subsection*{Prompt format for query completion and rule recovery tasks}

The large reasoning model (LRM) is prompted with the following structure for both the query completion and rule recovery tasks:

\paragraph{Section 1: Predicate definitions}
\textit{Here are some Predicate Definitions:}
\begin{itemize}
  \item \texttt{grandparent\_of(X,Y)}: X is a grandparent of Y. Order matters: the first argument is the grandparent, the second is the grandchild.
  \item \texttt{...} [Additional predicate definitions follow in the actual prompt]
\end{itemize}

\paragraph{Section 2: World rules}
There are three types of rules:
\begin{itemize}
  \item \textbf{A. Definite Rule:} Has a head and a body. It means if all atoms in the body are true, then the head is true.
  \item \textbf{B. Constraint:} Has only a body. It states that the atoms in the body cannot all be true at the same time.
  \item \textbf{C. Fact:} Has only a head. This atom is always true.
\end{itemize}

Variables are capitalized and rules with variables hold universally for all substitutions.

\textit{Here are the NoRA world rules. Rules are indexed and follow the format:}
\begin{quote}
\texttt{Head :- Body.}
\end{quote}
\vspace{0.3em}
\textit{Example:}
\begin{quote}
\texttt{1: grandparent\_of(Y,X) :- grandchild\_of(X,Y).}
\end{quote}
[All world rules are then enumerated by index.]

\paragraph{Section 3: Two exemplars}

\textbf{TASK:} You will be given a story made up of predicates describing relationships between entities ...

\textbf{Example 1:}
\begin{quote}
0 is a is\_person. \\
0 is a is\_female. \\
1 is a is\_place. \\
2 is a is\_person. \\
3 is a is\_person. \\
... [more story facts]
\end{quote}

\textit{Query:} What is the relation between 11 and 23? What are the indexes of the world rules you will need to derive this?

\textit{Response:}
\begin{itemize}
  \item \textbf{query\_label:} \texttt{niece\_of}
  \item \textbf{rules\_used:} \{\texttt{192, 64, 194, 46, 23}\}
  \item \textbf{reasoning:} \\
  From story fact 23, we know that individual 23 is the maternal aunt of individual 11. Applying world rule 192, we deduce that 23 is the maternal aunt or uncle of 11. Rule 194 generalizes this to \texttt{aunt\_or\_uncle\_of}. Rule 23 inverts this relation to yield that 11 is a \texttt{nibling} of 23. The story also indicates that 37 is the parent of 11 and has no sons. Applying rule 64, we infer that 11 is female. Rule 46 finally allows us to conclude that 11 is the \texttt{niece} of 23.
\end{itemize}

\textbf{Example 2:}
\begin{quote}
[Another guided exemplar with similar format]
\end{quote}

\paragraph{Section 4: Actual problem instance}

\textbf{STORY:}
\begin{quote}
0 is a is\_person. \\
1 is a is\_person. \\
1 is a is\_male. \\
2 is a is\_person. \\
... [More story facts]
\end{quote}

\textbf{QUERY:}
\begin{quote}
What is the predicate between 35 and 6? If a relationship between 35 and 6 is explicitly given in the story facts, and there is some other relationship that is also true, you need to uncover the unstated predicate. If multiple predicates capture the relationship between 35 and 6, choose the most specific one.

What are the indexes of the world rules you will need to derive this?
\end{quote}

\textbf{Expected Output:}
\begin{itemize}
  \item \textbf{query\_label:} \texttt{...}
  \item \textbf{rules\_used:} \{\texttt{...}\}
  \item \textbf{reasoning:} \texttt{...}
\end{itemize}

\section{Large reasoning models use shortcuts}\label{sec:LLRshortcuts}  
In the NoRA world rules, the knowledge that “a sibling of my sibling is also my sibling” is \emph{not} explicitly encoded as a definite rule.   
To prove it, one has to chain through the parent–child relations, repeatedly applying the following three world rules:

\begin{enumerate}[label=(W\arabic*)]
    \item \texttt{child\_of(Y,X) :- child\_of(Z\_1,X), sibling\_of(Y,Z\_1).}
    \item \texttt{parent\_of(Y,X) :- sibling\_of(Z\_1,X), parent\_of(Y,Z\_1).}
    \item \texttt{sibling\_of(Y,X) :- parent\_of(Z\_1,X), child\_of(Y,Z\_1), Y $\neq$ X.}
\end{enumerate}

Even before normalising gendered relations, establishing sibling transitivity therefore demands at least four inference steps.
%\paragraph{LLMs introduce an implicit shortcut.}
In contrast, LRMs 
%Pre-trained large language models (LLMs) 
have internalized the following direct rule:

\begin{enumerate}[label=(S)]
    \item \texttt{sibling\_of(X,Z) :- sibling\_of(X,Y), sibling\_of(Y,Z).}
\end{enumerate}

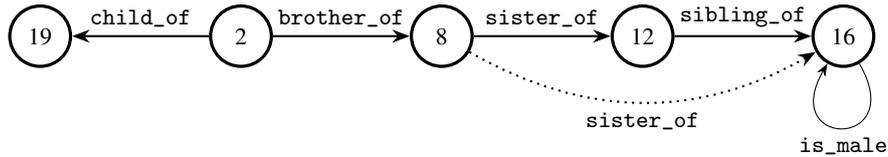
\begin{figure}[t]
    \centering
    \begin{tikzpicture}[
        node distance=1.8cm,
        rednode/.style={circle, fill=red!50, draw=red!80, very thick, minimum size=0.8cm},
        whitenode/.style={circle, fill=white, draw=black, very thick, minimum size=0.8cm},
        edge/.style={->, thick}
        ]
        \node[whitenode] (2) {2};
        \node[whitenode, left=of 2] (19) {19};
        \node[whitenode, right=of 2] (8) {8};
        \node[whitenode, right=of 8] (12) {12};
        \node[whitenode, right=of 12] (16) {16};

        \draw[edge] (2) -- node[midway, sloped, above] {\texttt{child\_of}} (19);
        \draw[edge] (2) -- node[midway, sloped, above] {\texttt{brother\_of}} (8);
        \draw[edge] (8) -- node[midway, above] {\texttt{sister\_of}} (12);
        \draw[edge] (12) -- node[midway, above] {\texttt{sibling\_of}} (16);
        \draw[->] (16) edge[out=300,in=240,looseness=8] node[below] {\texttt{\texttt{is\_male}}} (16);
        \draw[edge, dotted, bend right=30] (8) to node[midway, below] {\texttt{sister\_of}} (16);
    \end{tikzpicture}
    \caption{Illustrative fragment of the NoRA graph.  
    Solid edges follow world rules (W1–W3); the dotted edge shows the shortcut (S) inferred by the LRM. }
    \label{fig:example_graph}
\end{figure}

Rule (S) is not a NoRA world rule, yet LRMs (like o3) can apply it, collapsing a multi-hop proof into a single step.  
Consequently, these tasks that need high reasoning depth are effectively much shallower for such models. Every test instance that o3 solved at a reasoning depth $> 9$ contained sibling transitivity as a sub-problem, so the model’s actual reasoning depth was far lower than our theoretical estimate. An example instance with OPEC $>9$ that the o3 model predicts correctly is shown in \ref{fig:example_graph}.  

\section{Comparing BL and OPEC as measures of non-path reasoning}
\label{sec:bl_vs_opec}

The Backtrack Load (BL) is the ratio of the number of inference steps to the number of entities involved. As noted in Section~\ref{sec:DerSTeps}, the number of derivation steps is dependent on the way the world rules are set up. Since we have avoided including redundant rules when specifying the world rules, many problems have a large number of derivation steps. BL is therefore susceptible to overestimating non-path difficulty. An important advantage of BL, however, is that it is capable of identifying non-path reasoning even in cases without off-path edges. %, which makes up for the false positives.
On the other hand, OPEC can only identify non-path reasoning when there are off-path edges (i.e.\ edges which are not on any path between source and target), but it is not dependent on how the world rules are encoded. 
%It does not give false positives, but can give false negatives as it misses non-path reasoning on paths between source and target nodes.

BL and OPEC can be controlled independently. For the Test-D, Test-OPEC, and training datasets (as mentioned in the paper), we explicitly control these difficulty metrics to take values within certain limits. To investigate the true correlation between these two difficulty metrics, we explore the stories generated by ASP before sampling to curate datasets.

\begin{figure}[htbp]
    \centering
    \includegraphics[width=0.5\linewidth]{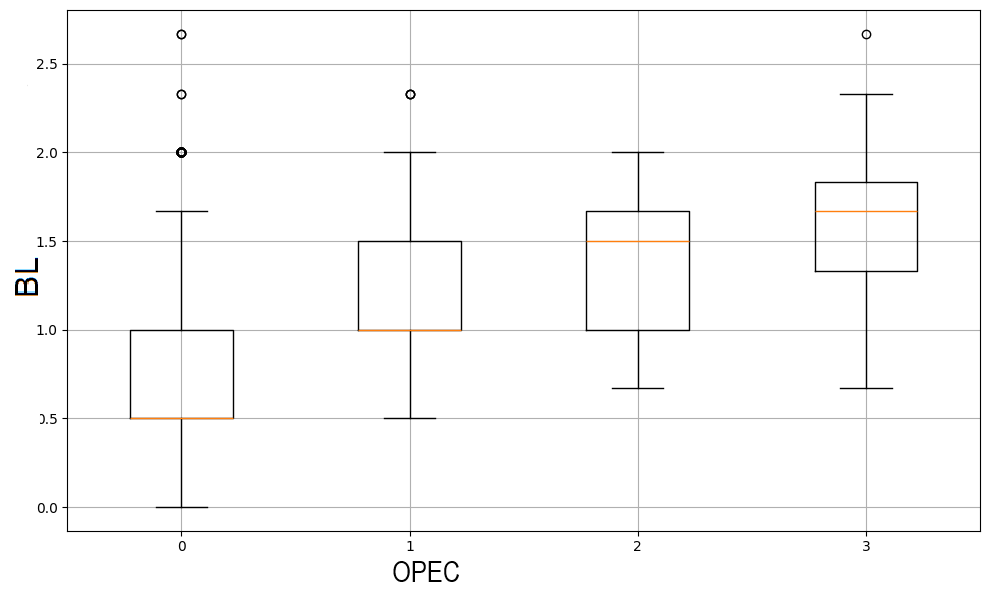}
    \includegraphics[width=0.5\linewidth]{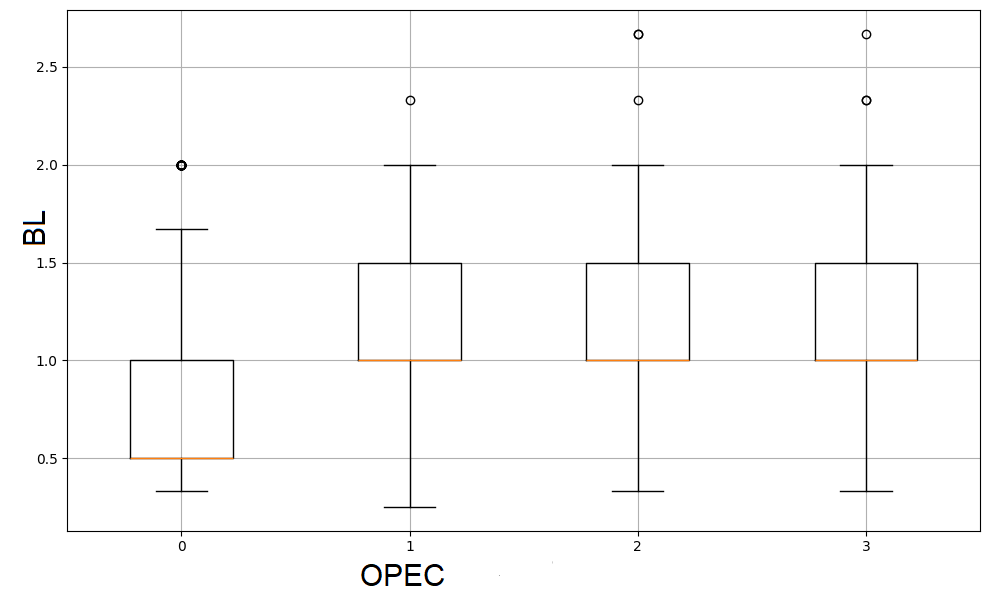}
    \caption{
    Illustration of the correlation between OPEC and BL using box plots of BL distributions for various OPEC values. 
    The top panel shows data generated \textbf{with ambiguous facts}, and the bottom panel shows data generated \textbf{without ambiguous facts}.
    }
    \label{fig:bl_opec_correlation}
\end{figure}

For the dataset with ambiguity, we observe a Pearson correlation coefficient between OPEC and BL of 0.321 (95\% confidence interval via bootstrap: [0.3086, 0.3359]). For the dataset without ambiguity, we observe a Pearson correlation coefficient between OPEC and BL of 0.4650 (95\% confidence interval via bootstrap: [0.4503, 0.4840]). Figure \ref{fig:bl_opec_correlation} breaks down this data.

\section{Ambiguous facts, story encodings and reasoning width}
\label{sec:AmbFacts} 

\begin{figure}[t]
	\centering
	
	% --- 2-a: Ambiguous Story Facts ---
	\begin{minipage}[t]{0.95\textwidth}
		\footnotesize
		\textbf{(a) Ambiguous Story Facts} \\
		\begin{tabular}{p{0.48\textwidth} p{0.48\textwidth}}
			\texttt{belongs\_to(ryan, underage).} & \textcolor{blue}{\texttt{1\{living\_in(cole, east\_rock);}} \\
			\texttt{school\_mates\_with(cole, will).} & \hspace{1em}\textcolor{blue}{\texttt{living\_in(cole, dwight)\}1.}} \\
			\texttt{living\_in\_same\_place(sheila, lalit).} & \textcolor{blue}{\texttt{1\{child\_of(ryan, brutus);}} \\
			\texttt{living\_in(lalit, kgp).} & \hspace{1em}\textcolor{blue}{\texttt{child\_of(ryan, cole)\}1.}} \\
			\texttt{living\_in(phil, kgp).} & \textcolor{blue}{\texttt{1\{colleague\_of(brutus, phil);}} \\
			& \hspace{1em}\textcolor{blue}{\texttt{colleague\_of(brutus, sheila)\}1.}} 
		\end{tabular}
	\end{minipage}
	
	\vspace{1.5em}
	 % --- 2-b: Reasoning Branches ---
	\begin{minipage}[t]{0.95\textwidth}		
		\textbf{(b) Refinements and Derivations} \\
		% \scriptsize
        \footnotesize
		% Row 1
		\begin{tabular}{p{0.48\textwidth} p{0.48\textwidth}}
			\begin{minipage}[t]{\linewidth}
            \strut\vspace*{-\baselineskip}\newline
				\colorbox{green!10}{\parbox{\linewidth}{
						\textbf{(i)} \\[-0.2em]
						\fbox{\parbox{0.96\linewidth}{
								\texttt{colleague\_of(brutus, phil), child\_of(ryan, brutus), living\_in(cole, east\_rock)}}} \\[0.4em]
						\texttt{living\_in(brutus, kgp) :- colleague\_of(brutus, phil), living\_in(phil, kgp).} \\[0.3em]
						\texttt{living\_in(ryan, kgp) :- belongs\_to(ryan, underage), parent\_of(brutus, ryan),} 
						\texttt{living\_in(brutus, kgp).}}}\\[1em]
                        \colorbox{green!10}{\parbox{\linewidth}{
						\textbf{(ii)} \\[-0.2em]
						\fbox{\parbox{0.96\linewidth}{
								\texttt{colleague\_of(brutus, phil), child\_of(ryan, brutus), living\_in(cole, dwight)}}} \\[0.4em]
						Same derivation as (i).}} \\[1em]

                        \colorbox{teal!10}{\parbox{\linewidth}{
						\textbf{(iii)} \\[-0.2em]
						\fbox{\parbox{0.96\linewidth}{
								\texttt{colleague\_of(brutus, sheila), child\_of(ryan, brutus), living\_in(cole, east\_rock)}}} \\[0.4em]
						\texttt{living\_in(sheila, kgp) :- living\_in\_same\_place(sheila, lalit), living\_in(lalit, kgp).} \\[0.3em]
						\texttt{living\_in(brutus, kgp) :- colleague\_of(brutus, sheila), living\_in(sheila, kgp).} \\[0.3em]
						\texttt{living\_in(ryan, kgp) :- belongs\_to(ryan, underage), parent\_of(brutus, ryan), living\_in(brutus, kgp).}}}                        	
			\end{minipage} &			
			\begin{minipage}[t]{\linewidth}
            \strut\vspace*{-\baselineskip}\newline
            \colorbox{teal!10}{\parbox{\linewidth}{
						\textbf{(iv)} \\[-0.2em]
						\fbox{\parbox{0.96\linewidth}{
								\texttt{colleague\_of(brutus, sheila), child\_of(ryan, brutus), living\_in(cole, dwight)}}} \\[0.4em]
						Same derivation as (iii).}}\\[1em]
                        
                    \colorbox{red!10}{\parbox{\linewidth}{
						\textbf{(v)} \\[-0.2em]
						\fbox{\parbox{0.96\linewidth}{
								\texttt{child\_of(ryan, cole), living\_in(cole, east\_rock),} \\
								\texttt{colleague\_of(brutus, sheila)}}} \\[0.4em]
						\texttt{belongs\_to(cole, underage) :- school\_mates\_with(cole, will).} \\[0.3em]
						\textcolor{red}{\texttt{:- belongs\_to(cole, underage), parent\_of(cole, ryan).}} \\
						Contradiction.}} \\[1em]

                	\colorbox{red!10}{\parbox{\linewidth}{
						\textbf{(vi)} \\[-0.2em]
						\fbox{\parbox{0.96\linewidth}{
								\texttt{child\_of(ryan, cole), living\_in(cole, dwight),} \\
								\texttt{colleague\_of(brutus, phil)}}} \\[0.4em]
						Same contradiction as (v).}}\\[1em]

                        \colorbox{red!10}{\parbox{\linewidth}{
						\textbf{(vii)} \\[-0.2em]
						\fbox{\parbox{0.96\linewidth}{
								\texttt{child\_of(ryan, cole), living\_in(cole, east\_rock),} \\
								\texttt{colleague\_of(brutus, phil)}}} \\[0.4em]
						Same contradiction as (v).}}\\[1em]

                        \colorbox{red!10}{\parbox{\linewidth}{
						\textbf{(viii)} \\[-0.2em]
						\fbox{\parbox{0.96\linewidth}{
								\texttt{child\_of(ryan, cole), living\_in(cole, dwight),} \\
								\texttt{colleague\_of(brutus, sheila)}}} \\[0.4em]
						Same contradiction as (v).}}
			\end{minipage}
		\end{tabular}
	\end{minipage}
	
	\caption{(a) An ambiguous story in NoRA, with three cardinality-based facts (highlighted in blue). (b) Each numbered box corresponds to a refinement. The top rectangle in each branch highlights the specific choices made for ambiguous facts, and the body shows the derivation of the entailed atom \texttt{living\_in(ryan, kgp)} or the contradiction that arises. }
	\label{fig:kanr_ambiguous_fig}
	\end{figure}

Real-world text is often ambiguous or incomplete. One motivation for including ambiguity in NoRA is that relation extraction pipelines based on coreference resolution can introduce noise or uncertainty. 
% Additionally, narratives themselves may be under-specified. Consider the story fragment: 
% \begin{quote}
% 	\emph{Paul went to his grandmother Sheila's house... Sheila's son Dixon was not happy with her decisions.}
% \end{quote}
% From this, it is unclear whether Dixon is Paul's father or uncle. 
To reflect such real-world uncertainty, NoRA includes ambiguous story-facts encoded in ASP using \emph{cardinality facts} of the form \texttt{l\{atom1; atom2; ...; atomk\}u}, which indicates that the number of true atoms in the set \{\texttt{atom1, atom2, ..., atomk}\} lies between \texttt{l} and \texttt{u} (both inclusive). 

Once such ambiguous facts are introduced into a story, the resulting logic program may admit multiple \emph{stable models}. An \textbf{entailed atom} in this setting is defined as an atom that is part of \emph{every} stable model but is not explicitly listed as a story fact. Figure~\ref{fig:kanr_ambiguous_fig}(a) shows an ambiguous story in NoRA that contains three ambiguous facts. These yield $2^3 = 8$ possible refinements, of which four result in contradictions, leaving four consistent stable models. A common atom across all four models is \texttt{living\_in(ryan, kgp)}, which is thus considered an entailed atom and may be used to construct  a dataset example instance. 

Ambiguity introduces a new notion of difficulty. For the entailed atom \texttt{living\_in(ryan, kgp)}, Figure~\ref{fig:kanr_ambiguous_fig}(b) shows eight refinements (i–viii), of which v–viii lead to contradictions and share the same structure. Among the positive refinements, refinement i and ii yield identical derivations, as do iii and iv. 
Intuitively, the \textbf{reasoning width} of a query is the sum of: 
\begin{itemize}
	\item the number of distinct derivations/proofs that yield the entailed atom across all stable models, and
	\item the number of distinct derivations/proofs that lead to contradiction in the remaining refinements.
\end{itemize}

For the example in Figure~\ref{fig:kanr_ambiguous_fig} (with story facts in 2a and the entailed atom \texttt{living\_in(ryan, kgp)}), this number is 3. A formal definition of \textbf{reasoning width} is provided in the main text.

In the stories of NoRA, only specific types of ambiguous facts are used. These follow the ASP cardinality format:

\begin{quote}
\texttt{l\{atom1; atom2; ...; atomk\}u}
\end{quote}

where $k \in\{ 2,3\}$, and either $u = l = 1$ (meaning \emph{exactly one} atom is true), or $l = 1$ and $u = k$ (meaning \emph{at least one} atom is true). The atoms used in such ambiguous facts are of the following types:

\begin{enumerate}
    \item \texttt{living\_in(a, b\textsubscript{i})}, where $a$ is a person and $b_i$ are different possible locations. The same $a$ appears in all atoms of the ambiguous fact, i.e., the ambiguity is over which location $a$ lives in.
    
    \item \texttt{rel(a, b\textsubscript{i})}, where $a$ and $b_i$ are persons, and \texttt{rel} is a binary predicate over people (e.g., \texttt{grandparent\_of}, \texttt{sibling\_of}). The same $a$ and the same \texttt{rel} are used in all atoms of the ambiguous fact, i.e., the ambiguity is over whom $a$ stands in relation to.
\end{enumerate}

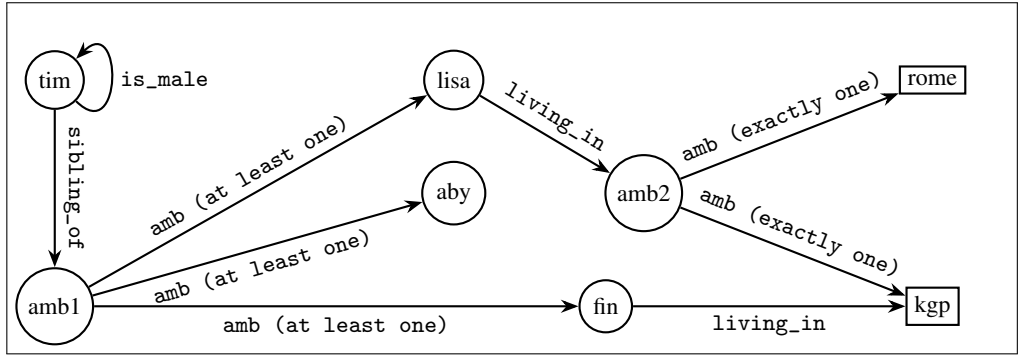
\begin{figure}[t]
\centering
\footnotesize

% \begin{tabular}{p{0.9\textwidth}}

% Panel (a) ASP
\begin{minipage}[t]{0.93\linewidth}
\textbf{(a) Story Facts in ASP Syntax} \\[1em]
\fbox{
    \begin{minipage}[t]{\linewidth}
    \texttt{1\{sibling\_of(tim,lisa); sibling\_of(tim,aby); sibling\_of(tim,fin)\}3}. \\
    \texttt{1\{living\_in(lisa,kgp); living\_in(lisa,rome)\}1}. \\
    \texttt{living\_in(fin,kgp)}.\\
    \texttt{belongs\_to\_group(tim,male)}.
    \end{minipage}
}
\end{minipage}

\vspace{20pt}

% Panel (b) Graph Encoding
\begin{minipage}[t]{0.93\linewidth}
\textbf{(b) Graph Encoding of Story Facts} \\[1em]
\fbox{\begin{minipage}[t]{\linewidth}
    \begin{tikzpicture}[->, thick, node distance=2cm, every node/.style={font=\footnotesize}]
        % Nodes
        \node[draw, circle] (tim) at (-.75,1.5) {tim};
        \node[draw, circle] (amb1) at (-.75,-1.5) {amb1};
        \node[draw, circle] (lisa) at (4.5,1.5) {lisa};
        \node[draw, circle] (aby) at (4.5,0) {aby};
        \node[draw, circle] (fin) at (6.5,-1.5) {fin};

        \node[draw, circle] (amb2) at (7,0) {amb2};
        \node[draw, rectangle] (kgp) at (10.8,-1.5) {kgp};
        \node[draw, rectangle] (rome) at (10.8,1.5) {rome};

        % Edges
        \draw[->] (tim) -- (amb1) node[midway, above,sloped] {\texttt{sibling\_of}};
        \draw[->] (amb1) -- (lisa) node[midway, above, sloped] {\texttt{amb (at least one)}};
        \draw[->] (amb1) -- (aby) node[midway, below, sloped] {\texttt{amb (at least one)}};
        \draw[->] (amb1) -- (fin) node[midway, below, sloped] {\texttt{amb (at least one)}};

        \draw[->] (lisa) -- (amb2) node[midway, above, sloped] {\texttt{living\_in}};
        \draw[->] (amb2) -- (kgp) node[midway, above, sloped] {\texttt{amb (exactly one)}};
        \draw[->] (amb2) -- (rome) node[midway, above, sloped] {\texttt{amb (exactly one)}};

        \draw[->] (fin) -- (kgp) node[midway, below] {\texttt{living\_in}};
        \draw[->] (tim) edge[out=-45,in=45,looseness=4] node[right] {\texttt{is\_male}} (tim);
    \end{tikzpicture}
    \end{minipage}
}
\end{minipage}

% \end{tabular}

\caption{(a) An example story in ASP syntax with two ambiguous facts. (b) Corresponding graph encoding: ambiguous facts are handled via auxiliary nodes with labeled ambiguity constraints on edges.}
\label{fig:amb_fact_encoding}
\end{figure}

\paragraph{Graph encoding of story facts.}  
When story facts are provided to a GNN, they must be converted into directed graphs. For non-ambiguous facts of the form \texttt{rel(a,b)}, we follow the standard convention: draw a directed edge from $a$ to $b$ with edge label \texttt{rel}. Special entities like \texttt{male} in relationships such as \texttt{belongs\_to\_group(sam,male)} are encoded as self-loops (e.g., \texttt{sam $\rightarrow$ sam} labeled \texttt{is\_male}), since neural models using these graphs rely solely on edge labels and cannot learn from node labels. %, translating ASP syntax into graph-theoretic format.

For ambiguous facts, a dedicated \textbf{ambiguous node} is introduced to maintain the structure and support model interpretability. Two types of constructions are used:

\begin{itemize}
    \item For ambiguous facts of the form \texttt{1\{rel(a,b\textsubscript{1}); rel(a,b\textsubscript{2}); ...; rel(a,b\textsubscript{k})\}k}, where \emph{at least one} relation is true:
    \begin{quote}
        Add an edge from $a$ to a newly created node $p\_amb\_node$ with label \texttt{rel}, and edges from $p\_amb\_node$ to each $b_i$ labeled \texttt{amb, at least 1 is true}.
    \end{quote}
    
    \item For ambiguous facts of the form \texttt{1\{rel(a,b\textsubscript{1}); rel(a,b\textsubscript{2}); ...; rel(a,b\textsubscript{k})\}1}, where \emph{exactly one} relation is true:
    \begin{quote}
        Add an edge from $a$ to $p\_amb\_node$ labeled \texttt{rel}, and edges from $p\_amb\_node$ to each $b_i$ labeled \texttt{amb, exactly 1 is true}.
    \end{quote}
\end{itemize}

Each ambiguous fact introduces exactly one such \texttt{p\_amb\_node}. This design allows GNN-based models to reason over ambiguous structures using only edge labels (see \ref{fig:amb_fact_encoding}). %, without requiring explicit node-level semantics. It supports structured disjunction reasoning in a graph format while preserving compatibility with conventional message-passing architectures.

\section{Diagnosing model performance on ambiguous stories}\label{appDiagnosingAppendixPerformance}

We initially expected that handling \emph{ambiguous stories} would pose a significant challenge for the models. However, to our surprise, most models performed nearly as well on the \texttt{Test-W} split as on the training splits. Upon further investigation, we identified that the metric we designed to measure reasoning difficulty---the \textbf{reasoning width}---does not fully capture some shortcuts that models can exploit to achieve high performance.

A problem instance $(S, a, b, R)$ with high reasoning width is difficult only if the solver adheres to the ideal reasoning process. However, models can take shortcuts that still very often lead to correct answers. For some instances, these shortcuts fail to yield correct predictions, and in such cases, the performance of the Edge Transformer model deteriorates substantially. Unfortunately, these challenging examples represent only a small fraction of the \texttt{Test-W} dataset.

\subsection{Illustrative example of ambiguity}

To illustrate this issue, consider the ambiguous story shown in Figure~\ref{fig:ambiguous_story}. This story contains one ambiguous fact---whether \texttt{Sean} or \texttt{Shah} is the colleague of \texttt{Rob}. This ambiguity gives rise to two refinements.

\begin{figure}[H]
    \centering
    \begin{tikzpicture}[
        every node/.style={font=\footnotesize},
        ->, thick, >=Stealth,
        baseline=(current bounding box.north)
    ]
        % Nodes
        \node[draw, circle] (rob) at (-1,2) {rob};
        \node[draw, rectangle] (amb1) at (2,2) {amb1};
        \node[draw, circle] (shah) at (5,1) {shah};
        \node[draw, circle] (sean) at (5,3) {sean};
        \node[draw, rectangle] (U) at (8,3) {U};
        \node[draw, rectangle] (V) at (8,1) {V};
        \node[draw, circle] (daisy) at (-4,2) {daisy};
        % Edges
        \draw[->] (rob) -- (amb1) node[midway, above] {\texttt{colleague\_of}};
        \draw[->] (amb1) -- (shah) node[midway, above, sloped] {\texttt{exactly\_one}};
        \draw[->] (amb1) -- (sean) node[midway, above, sloped] {\texttt{exactly\_one}};
        \draw[->] (sean) -- (U) node[midway, above] {\texttt{living\_in}};
        \draw[->] (shah) -- (V) node[midway, above] {\texttt{living\_in}};
        \draw[->] (shah) to [loop below] node[below] {\texttt{is\_underage}} (shah);
        \draw[->] (rob) to [loop above] node[above] {\texttt{is\_male}} (rob);
        \draw[->] (rob) -- (daisy) node[midway, above] {\texttt{sibling\_of}};
    \end{tikzpicture}
    \caption{Example ambiguous story containing one ambiguous fact: whether \texttt{Sean} or \texttt{Shah} is the colleague of \texttt{Rob}.}
    \label{fig:ambiguous_story}
\end{figure}
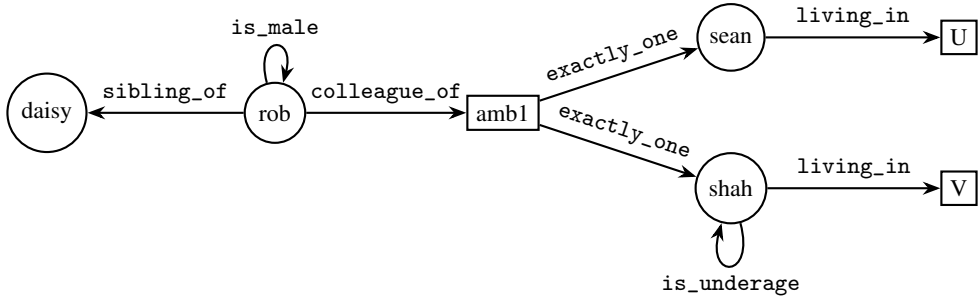

\begin{figure}[t]
\centering
\begin{minipage}{0.95\linewidth}
\footnotesize
\begin{multicols}{2}
\textbf{Query 1:} \texttt{rob is (brother, sibling) of daisy}

\vspace{0.04cm}
\begin{itemize}[leftmargin=*]
    \item \textbf{Refinement 1:} \{\texttt{sean is colleague of rob}\}  
    \textit{Derivation:} \texttt{rob} is a male sibling of \texttt{daisy} $\rightarrow$ \texttt{brother} relationship established.
    \item \textbf{Refinement 2:} \{\texttt{shah is colleague of rob}\}  
    \textit{Derivation:} \texttt{shah is underage} $\rightarrow$ cannot be colleague. \textcolor{red}{Contradiction. Negative refinement.}
\end{itemize}

\columnbreak

\textbf{Query 2:} \texttt{rob lives in U}

\vspace{0.01cm}
\begin{itemize}[leftmargin=*]
    \item \textbf{Refinement 1:} \{\texttt{sean is colleague of rob}\}  
    \textit{Derivation:} \texttt{sean lives in U} $\rightarrow$ colleague relationship suggests \texttt{rob} lives in \texttt{U}.
    \item \textbf{Refinement 2:} \{\texttt{shah is colleague of rob}\}  
    \textit{Derivation:} \texttt{shah is underage} $\rightarrow$ cannot be colleague. \textcolor{red}{Contradiction. Negative refinement.}
\end{itemize}
\end{multicols}
\end{minipage}
 \vspace{2mm}
\caption{Two queries derived from the ambiguous story, both with reasoning width 2. The first query can be solved even if the ambiguity is ignored; the second requires handling the ambiguity explicitly.}
\label{fig:two_queries}
\end{figure}

From this story, we derive two entailed facts that can each be turned into problem instances of reasoning width 2, shown in Figure~\ref{fig:two_queries}.
For the first query, an ideal reasoner would identify that the second refinement leads to a contradiction and reason accordingly, yielding two valid proofs—one per refinement—and thus a reasoning width of 2. However, a shortcut reasoner could ignore the ambiguous part of the story and solve the problem without learning and applying the contradiction rule that underage individuals cannot be colleagues. The second query also has reasoning width 2, but here it is essential to apply the contradiction and disprove the second refinement.

\subsection{Defining hard ambiguous instances in \texttt{Test-W}}

To obtain problem instances with ambiguity that are harder to solve, we devise an auxiliary criterion to distinguish the two types of ambiguous problem instances in \texttt{Test-W}.

\begin{definition}[Hard Ambiguous Problem Instances]
A problem instance in \texttt{Test-W} is labeled \textbf{hard} if:
\begin{enumerate}[label=(\roman*)]
    \item An ambiguous fact is used in the derivation of the entailed fact.
    % \item \textbf{All} possible resolutions of the choice rule \emph{fail} to derive the entailed fact.
    \item The entailed fact cannot be derived for all the possible resolutions of the ambiguous fact (i.e.\ some of the possible resolutions need to be excluded based on the fact that they violate the constraints).
\end{enumerate}
\label{def:hard_ambiguous}
\end{definition}

\begin{figure}[H]
\centering
\begin{tikzpicture}[
    every node/.style={font=\footnotesize},
    ->, thick, >=Stealth
]
    \node[draw,circle]   (sean)  at (-1,3) {sean};
    \node[draw,circle]   (daisy) at (-4,3) {daisy};
    \node[draw,rectangle](amb1)  at (2,3) {amb1};
    \node[draw,circle]   (lee)   at (5,4) {lee};
    \node[draw,circle]   (joe)   at (5,2) {joe};
    \draw (sean)  -- node[above]  {\texttt{parent\_of}}    (daisy);
    \draw (sean)  -- node[above] {\texttt{brother\_of}}   (amb1);
    \draw (amb1)  -- node[above,sloped] {\texttt{exactly\_one}} (lee);
    \draw (amb1)  -- node[above,sloped] {\texttt{exactly\_one}} (joe);
\end{tikzpicture}
 \vspace{2mm}
\caption{Story graph illustrating an example that satisfies condition (i) but not condition (ii) from Definition~\ref{def:hard_ambiguous}.}
\label{fig:story_graph_example}
\end{figure}
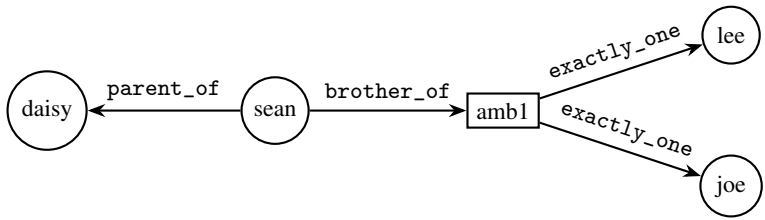

To illustrate condition (ii), consider the story graph in Figure~\ref{fig:story_graph_example}.

The entailed fact \texttt{father\_of(sean,daisy)} can be derived as follows:
\begin{itemize}[leftmargin=*]
    \item \textbf{Refinement 1:} From \texttt{brother\_of(sean,lee)} we derive \texttt{is\_male(sean)}
    \item \textbf{Refinement 2:} from \texttt{brother\_of(sean,joe)} we derive \texttt{is\_male(sean)}
\end{itemize}
For either resolution of the ambiguous fact we thus obtain \texttt{is\_male(sean)}. Together with \texttt{parent\_of(sean,daisy)} we thus derive \texttt{father\_of(sean,daisy)}.

This instance satisfies condition (i) but not condition (ii) of Definition~\ref{def:hard_ambiguous}.

\subsection{Performance on hard ambiguous instances}

The truly challenging examples in \texttt{Test-W} (i.e., those satisfying Definition~\ref{def:hard_ambiguous}) are rare. However, for these examples, the accuracy of the Edge Transformer is substantially lower, as shown in Table~\ref{tab:hard_instances}.

\begin{table}[t]
    \centering
    \caption{Edge Transformer performance on \texttt{Test-W} subsets. For comparison, the in-distribution accuracy (same difficulty metric as training) is 90\%.}
    \label{tab:hard_instances}
    \begin{tabular}{lrr}
        \toprule
        & \# Examples & Exact Match Accuracy \\
        \midrule
        \textbf{Hard Ambiguous Instances} & 390  & 51\% \\
        \textbf{Non-Hard} & 6062 & 81\% \\
        \bottomrule
    \end{tabular}    
\end{table}

\section{NoRA v1.1}
\label{sec:NoRA1.1}

The world rules for \textsc{NoRA-1.1} are largely the same as those used in \textsc{NoRA}. We introduced a few targeted adjustments to address shortcut behaviors identified in Appendix~\ref{sec:LLRshortcuts}. In addition, \textsc{NoRA-1.1} does \emph{not} include stories with ambiguity. For reference, the complete world-rule specifications for \textsc{NoRA-1.1}, \textsc{NoRA}, and \textsc{InspiredFromHetionet} are available at:
\url{https://github.com/axd353/WhenNoPathsLeadToRome/tree/main/ExplicitWorldRuleFilesForReference}.

\begin{figure}[H]
  \centering
  \begin{subfigure}{0.95\linewidth}
    \centering
    \includegraphics[width=\linewidth]{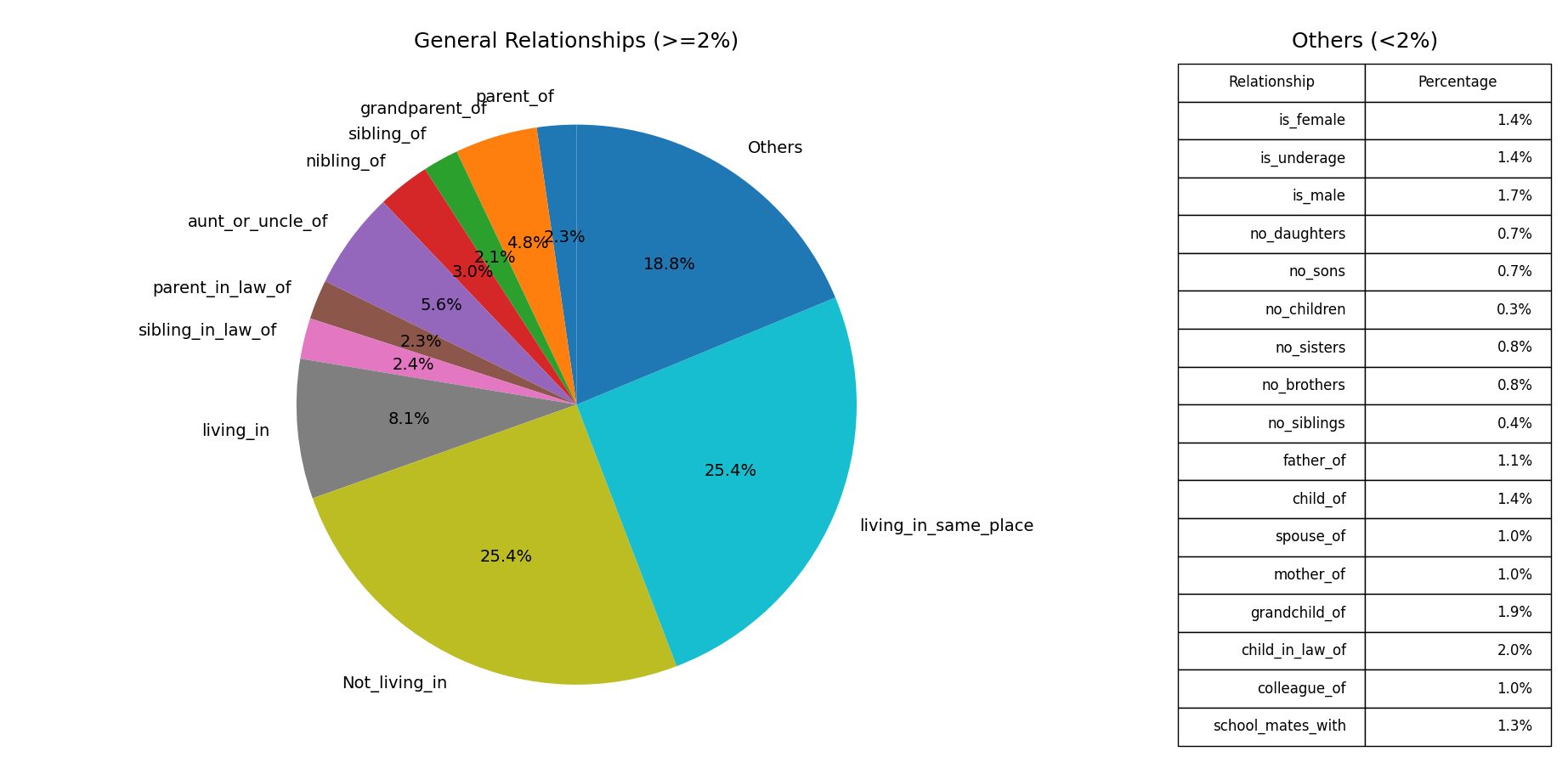}
    \caption{Distribution of predicates/relationships in the \textsc{NoRA-1.1} training set.}
    \label{fig:nora11_pred_dist}
  \end{subfigure}

  \vspace{0.75em}

  \begin{subfigure}{0.95\linewidth}
    \centering
    \includegraphics[width=\linewidth]{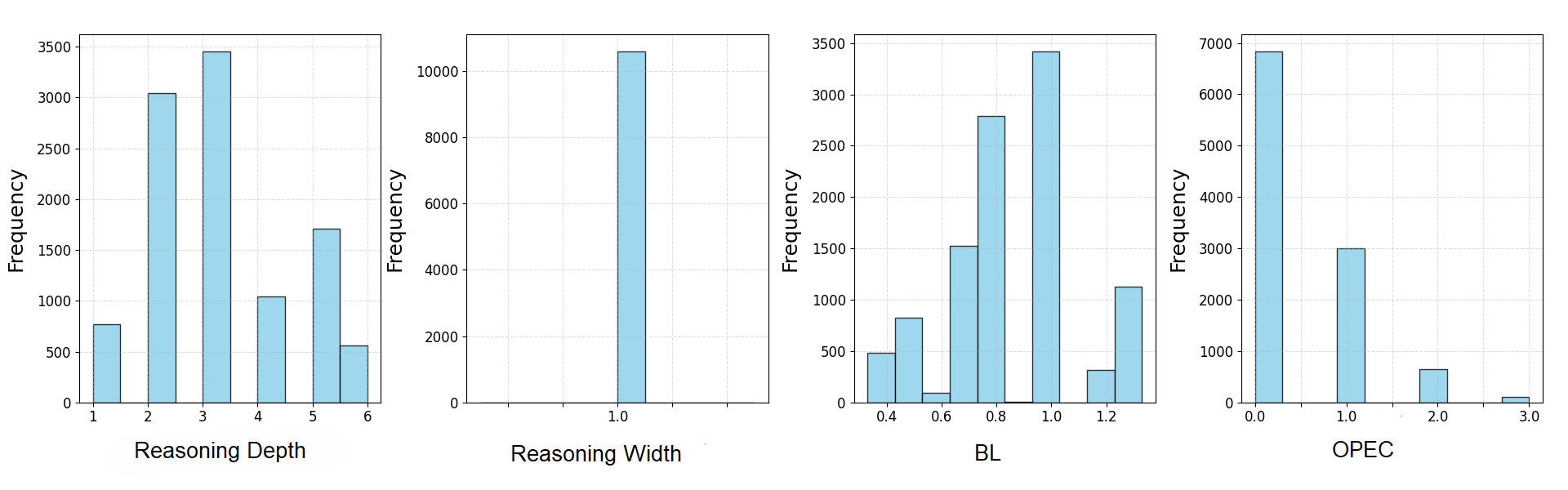}
    \caption{Distribution of difficulty metrics for problem instances in the \textsc{NoRA-1.1} training split.}
    \label{fig:nora11_difficulty_dist}
  \end{subfigure}

  \caption{\textsc{NoRA-1.1} training-set statistics. Top: predicate/relationship frequencies. Bottom: difficulty-metric distribution for training instances.}
  \label{fig:nora11_train_stats}
\end{figure}

The training split and the various test splits for \textsc{NoRA-1.1} are organized as shown in Table~\ref{tab:nora11_test_sets}. Figure \ref{fig:nora11_train_stats} shows some basic statistics of the NoRA v1.1 training set.

\begin{table}[t]
\centering
\caption{Overview of the NoRA v1.1 dataset splits. Values that require generalization beyond the training distribution are highlighted in red.}
\label{tab:nora11_test_sets}
\footnotesize
\setlength\tabcolsep{6pt} % wider since we have full width
\begin{tabular}{lcccc}
\toprule
\textbf{Name}  &  \textbf{Depth} &  \textbf{Width} &  \textbf{BL}  &  \textbf{OPEC} \\
\midrule
\textbf{Train-\blue{na}}                 & $\leq 6$ & \textbf{\blue{1}} & $< 1.5$ & $\leq 3$ \\
\midrule
\textbf{Test-\red{D}-\blue{na}}          & \red{$\mathbf{> 6}$} & \textbf{\blue{1}} & $< 1.5$ & $\leq 3$ \\
\textbf{Test-\red{BL}-\blue{na}}         & $\leq 6$ & \textbf{\blue{1}} & \red{$\mathbf{ \geq 1.5}$} & $\leq 3$ \\
\textbf{Test-\red{OPEC}-\blue{na}}       & -- & \textbf{\blue{1}} & -- & \red{$\mathbf{\geq 3}$} \\
\textbf{Test-In-dist-\blue{na}}          & $\leq 6$ & \textbf{\blue{1}} & $< 1.5$ & $\leq 3$ \\
\bottomrule
\end{tabular}
\end{table}

\section{Derivation step sensitivity}
\label{sec:DerSTeps} 
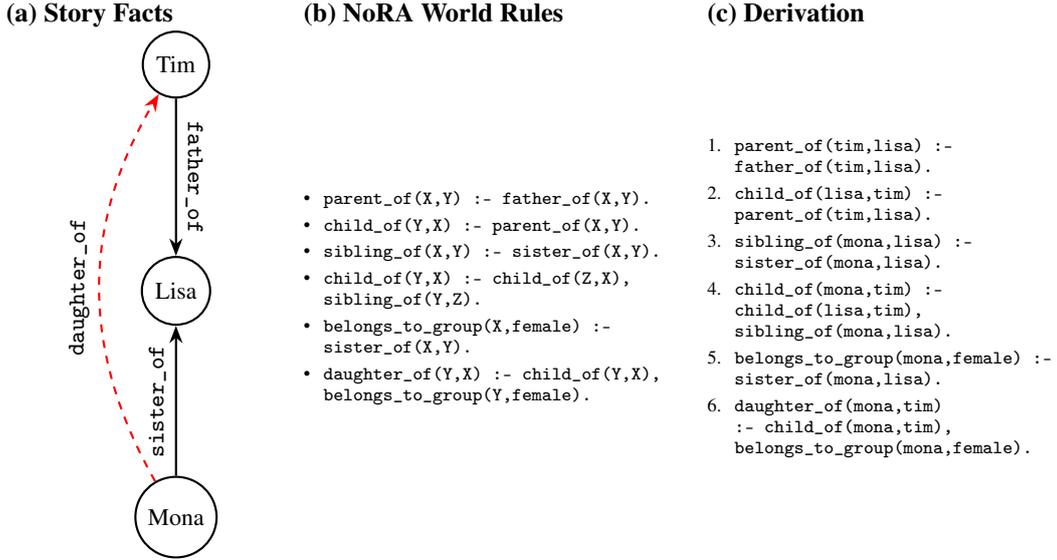
\begin{figure}[h]
	\centering
	\begin{tabular}{m{0.25\textwidth}m{0.35\textwidth}m{0.35\textwidth}}
		\textbf{(a) Story Facts} & \textbf{(b) NoRA World Rules} & \textbf{(c) Derivation} \\
		\begin{minipage}[t][5cm]{\linewidth}
			\centering
        \begin{tikzpicture}[node distance=2cm, ->, thick, every node/.style={font=\footnotesize}]
	\node[draw, circle] (tim) at (0,0) {Tim};
	\node[draw, circle] (lisa) at (0,-3) {Lisa};
	\node[draw, circle] (mona) at (0,-6) {Mona};
	
	\draw[->] (tim) -- (lisa) node[sloped, midway, above] {\texttt{father\_of}};
	\draw[->] (mona) -- (lisa) node[sloped, midway, above] {\texttt{sister\_of}};
	\draw[->, red, dashed] (mona) to [bend left=30] node[sloped, midway, above, black] {\texttt{daughter\_of}} (tim);
\end{tikzpicture}
		\end{minipage}
		&
		\scriptsize
		\begin{itemize}[leftmargin=*,topsep=0pt,itemsep=0pt]
			\item \texttt{parent\_of(X,Y) :- father\_of(X,Y).}
			\item \texttt{child\_of(Y,X) :- parent\_of(X,Y).}
			\item \texttt{sibling\_of(X,Y) :- sister\_of(X,Y).}
			\item \texttt{child\_of(Y,X) :- child\_of(Z,X), sibling\_of(Y,Z).}
			\item \texttt{belongs\_to\_group(X,female) :- sister\_of(X,Y).}
			\item \texttt{daughter\_of(Y,X) :- child\_of(Y,X), belongs\_to\_group(Y,female).}
		\end{itemize}
		&
		\scriptsize
		\begin{enumerate}[leftmargin=*,topsep=0pt,itemsep=0pt]
			\item \texttt{parent\_of(tim,lisa) :- father\_of(tim,lisa).}
			\item \texttt{child\_of(lisa,tim) :- parent\_of(tim,lisa).}
			\item \texttt{sibling\_of(mona,lisa) :- sister\_of(mona,lisa).}
			\item \texttt{child\_of(mona,tim) :- child\_of(lisa,tim), sibling\_of(mona,lisa).}
			\item \texttt{belongs\_to\_group(mona,female) :- sister\_of(mona,lisa).}
			\item \texttt{daughter\_of(mona,tim) :- child\_of(mona,tim), belongs\_to\_group(mona,female).}
		\end{enumerate}
	\end{tabular}
	\caption{Example showing a derivation using a minimal number of rules.}\label{fig:illustrateNORA}
\end{figure}
The number of derivation steps is sensitive to the precise way in which world rules are framed. To illustrate this, consider our NoRA world rules, which are designed to be minimal and avoid redundancy. These rules imply certain relationships (which are not explicit in the world rules). A model could also memorize these implied rules. This would result in shorter derivations but would necessitate memorizing a larger number of rules.

Consider the example in Figure \ref{fig:illustrateNORA}. Using the NoRA rules, entailing that Mona is the daughter of Tim requires six derivation steps. However, a rule not explicitly stated in the NoRA world rules, but implied by them, is:
\begin{align*}
\texttt{daughter\_of(Z,Y) :- father\_of(X,Y), sister\_of(Z,Y).}
\end{align*}
If models were to learn such implied rules directly, the derivation for the same entailment would be reduced to a single step. 

CLUTRR does not count inverse relationships, such as $\texttt{parent\_of(X,Y) :- child\_of(Y,X)}$, as derivation steps, whereas such steps are counted in NoRA. Since we have diverse types of rules in NoRA, making a judgment on what counts as a derivation step requires more consideration. 
%On the other hand, the purpose of NoRA is to test problem difficulty in the ambiguity and non-pathness direction; the number of steps is more aligned to reasoning depth, which has been tested as a source of difficulty extensively in CLUTTR \cite{Sinha2019CLUTRR}.

\section{Stable models}
\label{sec:stab_mod}
Solving a logic program  involves computing its \textbf{stable models}, which are also known as \textbf{answer sets} \citep{lifschitz2008twelve}. 
First note that while we usually specify ASP programs using rules with variables, the semantics of answer sets is defined w.r.t.\ the grounding of such programs. A ground rule is obtained by replacing the variables in an ASP rule by constants that appear in the program. The grounding of an ASP program consists of all the possible ground rules that we can obtain from its rules. Let us now assume that $P$ is a ground program (i.e.\ the grounding of an ASP program).

In the absence of rules without negation-as-failure, a stable model of $P$ is a minimal set of atoms, such that:
\begin{enumerate}
	\item If we assign \texttt{true} to every atom in the set, and \texttt{false} to all other possible atoms, then all rules in $P$ are satisfied.
	\item No strict subset of the model satisfies the above condition.
\end{enumerate}
For rules with negation-as-failure, answer sets are defined in terms of the Gelfond-Lifchitz reduct. While we do not explicitly rely on negation-as-failure in our encoding, for ambiguous facts (see below), we use a language construct that under the hood is translated to such rules. Some of the rules then have conditions with negation-as-failure, of the form $\textit{not}\, r(a,b)$. Such conditions are intuitively satisfied unless $r(a,b)$ can be inferred. The Gelfond-Lifschitz reduct of a logic program $P$ w.r.t.\ the answer set $A$ is the logic program $P^A$ that we obtain as follows:
\begin{itemize}
\item Any rule with a condition of the form $\textit{not}\, r(a,b)$ such that $r(a,b)\in A$ is removed from the program.
\item Every condition of the form $\textit{not}\,r(a,b)$ such that $r(a,b)\notin A$ is removed from the body of the rule in which it occurs.
\end{itemize}
Note that the reduct $P^A$ no longer contains negation-as-failure.
We then say that $A$ is an answer set of $P$ iff it is an answer set of the reduct $P^A$.

Intuitively, a stable model includes both the explicitly stated story facts and additional atoms that follow logically.

\end{document}